\ifdefined\pdfobjcompresslevel
\fi
\RequirePackage{fix-cm}

\RequirePackage{scrlfile}
\PreventPackageFromLoading{CJKutf8}
\PassOptionsToPackage{table,dvipsnames}{xcolor}
\PassOptionsToPackage{sort&compress}{natbib}
\documentclass[11pt,a4paper]{jitagent}
\usepackage{xeCJK}

\setCJKsansfont{FandolHei-Regular.otf}[BoldFont=FandolHei-Bold.otf]
\setCJKmonofont{FandolFang-Regular.otf}
\usepackage{graphicx,booktabs,tabularx,array,multirow}
\usepackage{placeins}
\usepackage[noabbrev,nameinlink]{cleveref}
\usepackage{enumitem}
\usepackage{fontawesome5}
\usepackage{atria-jit-branding}
\usepackage{tikz}
\usetikzlibrary{arrows.meta,positioning,calc,fit,backgrounds}
\definecolor{atria}{HTML}{0E419C}
\definecolor{atrialight}{HTML}{F0F4FB}
\definecolor{atriablue}{HTML}{315F9C}
\definecolor{atriagold}{HTML}{9B6B25}
\hypersetup{colorlinks=true,linkcolor=atria,citecolor=atriablue,urlcolor=atriablue,
  pdfauthor={CoworkOS Team},
  pdftitle={CoworkOS: Scaling Agent Coworkers for Real-world Automation}}
\setlist[itemize]{leftmargin=1.3em,itemsep=2pt,topsep=4pt}
\setlist[enumerate]{leftmargin=1.5em,itemsep=2pt,topsep=4pt}
\newcolumntype{Y}{>{\raggedright\arraybackslash}X}
\newcommand{\system}{\textsc{CoworkOS}}

\posttitle{\par\endgroup\vspace{0.55em}}
\preauthor{\par\begingroup\centering\sffamily\bfseries\fontsize{15pt}{18pt}\selectfont}
\postauthor{\par\endgroup\vspace{0.40em}}

\title{\hspace*{0.35em}\raisebox{-6.5pt}{\includegraphics[height=1.28em]{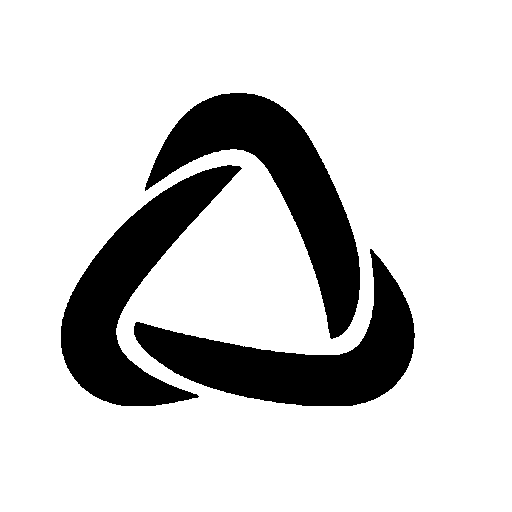}}\hspace{0.22em}CoworkOS: Scaling Agent Coworkers~for\\[-0.08em] Real-world Automation}

\author{\hyperref[sec:author-contributions]{CoworkOS Team}}

\renewcommand{\institutionlogos}{%
  \raisebox{3em}{\includegraphics[height=1.5cm]{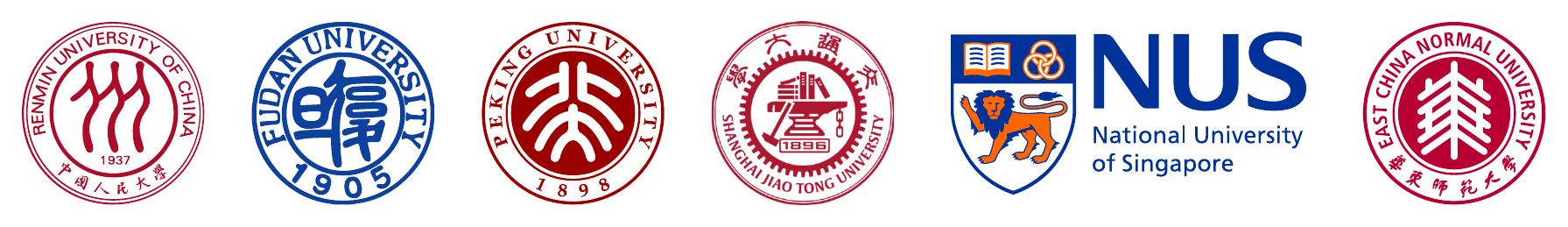}}%
}
\renewcommand{\institutionlogos}{%
  \includegraphics[height=1.4cm]{fig/institution-logos-strip.pdf}%
}

\renewcommand{\placeinstitutionheader}{%
  \AddToShipoutPictureFG*{%
    \put(\LenToUnit{\dimexpr1in+\oddsidemargin\relax},%
         \LenToUnit{\dimexpr\paperheight-1.2cm+0.9em\relax}){%
      \makebox(0,0)[lt]{\institutionlogos}}%
    \put(\LenToUnit{\dimexpr1in+\oddsidemargin\relax},%
         \LenToUnit{\dimexpr\paperheight-2.5cm\relax}){%
      \makebox(0,0)[lt]{{\normalcolor\rule{\textwidth}{0.4pt}}}}%
  }%
}

\metadata[\textcolor{atria}{\faGlobe}\ Project Page]{\href{https://coworkos-ai.github.io/}{\texttt{coworkos-ai.github.io}}}
\metadata[\textcolor{atriablue}{\faGithub}\ Repo Link]{\href{https://github.com/RUC-NLPIR/CoworkOS}{\texttt{github.com/RUC-NLPIR/CoworkOS}}}
\metadata[\textcolor{atriablue}{\faImages}\ Demo Gallery]{\href{https://coworkos-ai.github.io/showcase.html}{\texttt{coworkos-ai.github.io/showcase.html}}}
\metadata[\textcolor{atriablue}{\faEnvelope}\ Contact]{\email{jinjiajie@ruc.edu.cn}, \email{tgui@fudan.edu.cn}, \email{dou@ruc.edu.cn}}

\begin{document}
\placeinstitutionheader
\maketitle
\vspace{0em}
\begin{abstract}
Rapid advances in cowork agents are shifting human-AI collaboration toward real-world automation, delivering software, designs, and other artifacts. However, such productivity work demands both high-quality delivery on complex productivity tasks and continued monitoring and decision-making over long horizons, a combination that remains an open challenge. We introduce \system{}, a general-purpose harness for long-horizon work that separates task strategy from execution management: the model decides actions, delegation, and acceptance, while the runtime maintains state and feedback. Its three-layer architecture enables direct execution or delegation to Subagents, synchronously or asynchronously, under stable responsibilities. Through a shared application layer, humans and agents inspect and revise the same evolving documents, presentations, CAD models, and Blender scenes, so human-in-the-loop participation can enter both through conversation and through direct modification of the work itself. To sustain continuity across agents and sessions, \system{} further introduces hierarchical workspaces and persistent semantic memory, so working files, key decisions, and long-running Job logs remain shareable, reusable, and resumable after interruption. Across 12 benchmarks and 6 categories of real-world applications, \system{} proves highly competitive with frontier agent applications such as Codex and DeepSeek Harness. As a sustained-work case study, \system{} uses GLM-5.3 as its backbone to conduct RSI-driven automated post-training on Qwen3.6-35B-A3B. Over five rounds, the model's average score across eight benchmarks rose from 33.19 to 42.95 (+9.76). Looking ahead, we believe \system{} opens a meaningful path toward general cowork-agent intelligence.

\printprojectmetadata
\end{abstract}
\begingroup
\setlength{\intextsep}{6pt}
\begin{figure}[H]
\centering
\includegraphics[width=\linewidth]{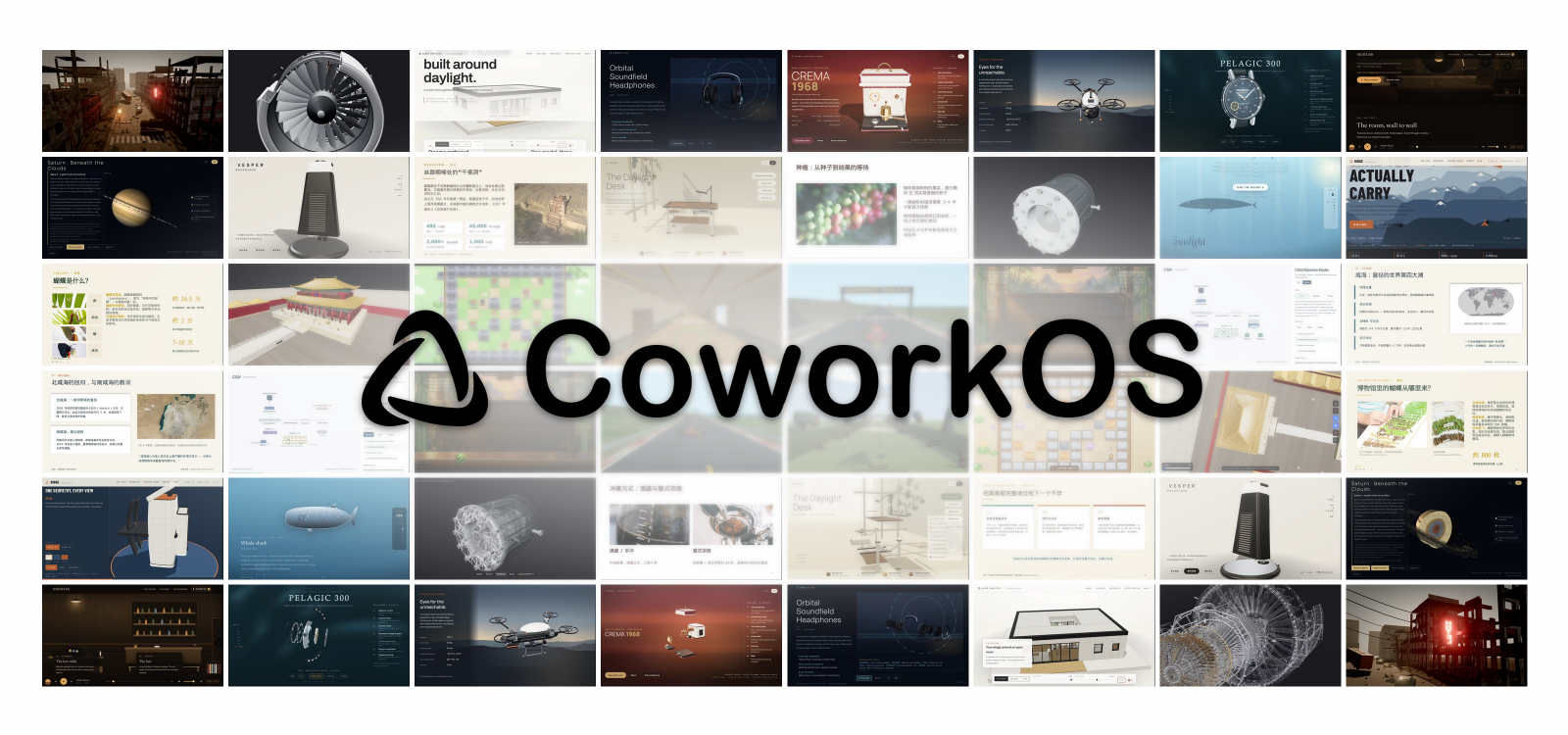}
\caption{Overview of 24 application cases produced with \system{} across six categories: frontend websites, games, Office presentations, interactive visualizations, CAD, and Blender rendering. Two views of each case illustrate different pages, scenes, or inspection states. The montage provides a visual overview; harness comparisons and detailed inspection evidence follow in Figure~\ref{fig:four-category-comparison} and Appendix~\ref{sec:case-protocols}.}
\label{fig:showcase-overview}
\end{figure}
\endgroup

\section{Introduction}
\label{sec:introduction}

Agent intelligence is advancing toward lifelong autonomy systems that continually interact and evolve with real-world environments. Progress toward Artificial Super Intelligence (ASI) requires attention to both model capability and the runtime systems that translate reasoning into action. Ideally, such agents would invoke tools, interpret feedback, maintain working state, and continue when their immediate interaction ends. To achieve this, the \emph{agent harness} supplies the runtime mechanisms for this process, including tool orchestration, state and context management, and execution feedback. Because the harness shapes how a model acts and how well it can use available resources, models and their harnesses need to co-evolve~\citep{liu2024agentbench,sumers2024cognitive}.

Recently, the evolution of agent systems can be understood through three progressive stages. \textbf{Agent Workflow} centers on reasoning and tool-using trajectories: early mechanisms include Chain-of-Thought and Tree-of-Thoughts~\citep{wei2022cot,yao2023tot}, interleaved reasoning and action in ReAct~\citep{yao2023react}, and feedback-driven reflection in Reflexion~\citep{shinn2023reflexion}. Workflow frameworks and construction methods, including DSPy~\citep{khattab2023dspy}, AFlow~\citep{zhang2024aflow}, and MetaGPT~\citep{hong2024metagpt}, further organize these capabilities. \textbf{Agent Assistant} extends this scope to persistent individual needs, bringing memory, messaging, code execution, and tool interfaces into continuing interactions. Frameworks such as LangChain, LangGraph~\citep{langchain2026langgraph}, AutoGen~\citep{wu2023autogen}, and OpenClaw support different aspects of this development. The emerging \textbf{Agent Cowork} setting places agents within professional work: people and agents contribute to evolving artifacts, adjust their plans as work develops, and share initiative in complex productivity tasks. Coding and work-environment agents represented by Claude Code, Codex, OpenHands, and DeepSeek Harness have demonstrated strong real-world productivity~\citep{anthropic2026claudecode,deepseek2026harness,openai2026codex,wang2025openhands,jimenez2024swebench}, while cowork agents such as Kimi Code, ZCode, Qoder, and Hermes Agent have also been widely adopted. Notably, these stages describe an expanding design lens: from automation by the agent itself, gradually toward sustained interactive work involving humans, agents, and the environment~\citep{liu2024agentbench,xie2024osworld}.

\begin{figure}[!htbp]\centering
\includegraphics[width=\linewidth]{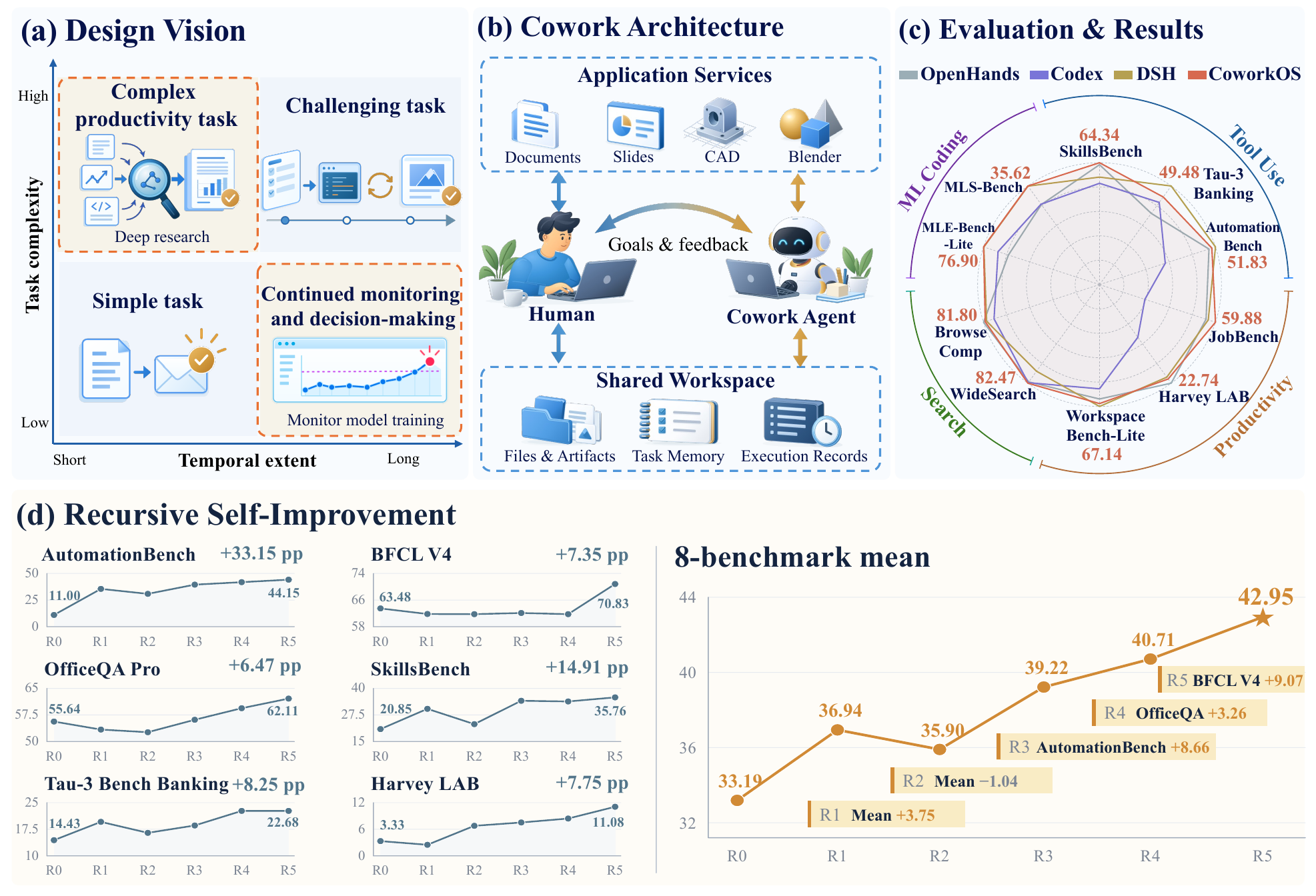}
\caption{Overview of \system{}. \textbf{(a)} Two complementary requirements for cowork agents: complex productivity tasks and continued monitoring and decision-making. \textbf{(b)} Human--agent cowork through shared, evolving work. \textbf{(c)} Performance overview of \system{} alongside other frontier cowork harnesses across 10 benchmarks in four domains. \textbf{(d)} Recursive self-improvement (RSI) via automated post-training over five improvement rounds. Details are provided in Section~\ref{sec:rsi} and Appendix~\ref{sec:rsi-provenance}.}
\label{fig:intro-overview}
\end{figure}

Building on this insight, we argue that this expansion of productivity reveals two complementary requirements for today's cowork agents, illustrated in Figure~\ref{fig:intro-overview}(a):
\begin{itemize}
  \item \textbf{Complex productivity tasks:} producing software, documents, designs, and other artifacts through demanding reasoning, tool use, and verification. Such tasks call for the agent's comprehensive agentic intelligence and, where the work requires it, collective intelligence across collaborating agents, to carry a difficult task through to completion.
  \item \textbf{Continued monitoring and decision-making:} monitoring a training process over days, following a project across sessions, or watching external information and deciding whether and how to act when a relevant event occurs. Such monitoring requires the agent to keep its context healthy over time, maintain sound memory, and continuously manage and track the artifacts it produces.
\end{itemize}
Notably, both settings require sustained collaboration between humans and agents, which we treat as a basic design principle for cowork agents: the two sides work together within a shared application ecosystem centered on evolving work (Figure~\ref{fig:intro-overview}(b)). This leads to our central question: can a single cowork agent handle both complex productivity tasks and continued monitoring and decision-making, while keeping humans in the loop?

We propose \system{}, a cowork harness for long-horizon and human-in-the-loop task execution. To achieve this, a natural approach is a division of responsibility: \emph{the model autonomously chooses the task-execution strategy, while the runtime harness maintains efficient, long-horizon cowork mechanisms}. Specifically, \system{} consists of a \textbf{three-layer agent framework} comprising a Main Agent, a Coordinator Runtime, and optional Subagents. Main sets the overall task plan and execution direction, decomposes work as needed, and evaluates the final artifacts and results returned. The Coordinator dispatches subtasks to the corresponding agents and monitors their lifecycles. Subagents carry out their assignments within the context and resources provided to them, using tools and making local decisions. Thanks to this hierarchy, \system{} can support complex productivity tasks through collective intelligence, while the Coordinator enables Subagents to sustain continued monitoring and decision-making on long-horizon tasks.

To realize this collaboration pattern at scale, we build a \textbf{human-in-the-loop scalable application ecosystem}. It provides foundational tool capabilities including CLI execution, search, file operations, and MCP interfaces, together with extensible access to professional applications, including document editors, PPTist, CAD, and Blender. These tools support work across code, office artifacts, three-dimensional content, and games. A distinctive feature is the shared application object: people and agents can inspect and revise the same evolving document, model, or scene. Domain operations, rendered views, and other supported feedback inform subsequent decisions, so human participation can enter both through conversation and through direct modification of the work and its artifacts, embedding human-in-the-loop participation within autonomous execution.

To sustain persistent memory and communication across agents, we introduce \textbf{hierarchical workspaces and persistent semantic memory}, which provide continuity for the task. Shared workspaces let agents inspect and reuse one another's working files and candidate artifacts; semantic memory records key decisions, constraints, and unfinished work; and long-running operations run as Jobs that retain their status and logs, so work can be picked up accurately after waiting or interruption.

As shown in Figure~\ref{fig:intro-overview}(c), we evaluate \system{} on 12 general agent benchmarks and across 6 categories of real-world application scenarios. \system{} shows competitive performance relative to strong mainstream cowork frameworks such as OpenHands, Codex, and DeepSeek Harness.

To further explore the potential of \system{}, we task it with \textbf{RSI-native automated model post-training}, a workload that naturally combines complex productivity work with continued monitoring and decision-making. Specifically, \system{} automatically analyzes model weaknesses, optimizes the data pipeline, synthesizes and validates post-training data, feeds it into training, and iterates this closed loop. Doing so requires the model not only to decide wisely how the data pipeline should be optimized, but also to monitor long-running training runs and preserve their execution logs. Over five improvement rounds, \system{} raises the average score of Qwen3.6-35B-A3B from 33.19 to 42.95 across eight challenging productivity benchmarks, a gain of 9.76 points, making strategic optimization decisions throughout the iterations and demonstrating its strong potential as an RSI-native coworker. Qualitative comparisons and post-training run records are detailed in Section~\ref{sec:cases} and Appendix~\ref{sec:eval-details}.

This report makes four contributions:
\begin{itemize}
  \item \textbf{A cowork harness architecture combining complex productivity tasks with continued monitoring and decision-making.} A division of responsibility lets the model autonomously choose the task-execution strategy while the runtime harness maintains efficient, long-horizon cowork mechanisms; the three-layer framework of Main Agent, Coordinator Runtime, and Subagents supports complex productivity tasks through collective intelligence and sustains continued monitoring and decision-making on long-horizon tasks.
  \item \textbf{A human-in-the-loop scalable application ecosystem.} Foundational tools and extensible professional applications including documents, presentations, CAD, and Blender support code, office artifacts, three-dimensional content, and games; shared, evolving application objects let humans participate through conversation or by directly modifying the work itself, embedding human-in-the-loop participation within autonomous execution.
  \item \textbf{Persistent working state and execution continuity.} Hierarchical workspaces and persistent semantic memory keep working files, artifacts, decisions, and unfinished work shareable and reusable across agents, while Jobs retain the status and logs of long-running operations so work can resume accurately after waiting or interruption.
  \item \textbf{Cross-domain generalization with RSI-native capability.} Across 12 benchmarks in four domains and six categories of real-world cowork scenarios, \system{} generalizes across tool use, productivity, search, and ML coding with performance comparable to strong mainstream cowork frameworks; in a five-round RSI-native post-training study, it further raises the average score of Qwen3.6-35B-A3B from 33.19 to 42.95 over eight benchmarks.
\end{itemize}


\section{Preliminaries: Task Formulation and Agency}
\label{sec:preliminaries}

We first define the tasks an agent system is expected to solve, then describe our view of agency and two dimensions that characterize the challenges these tasks present.

\subsection{General Agent Task Formulation}
\label{sec:formulation}

An agent task is specified by
\begin{equation}
  \mathcal T=(g,\mathcal E,\mathcal A,\mathcal V,B),
  \label{eq:task-specification}
\end{equation}
where $g$ is the user's objective, $\mathcal E$ the environment, $\mathcal A$ the action space, $\mathcal V$ the acceptance criterion, and $B$ the resource budget. The environment can include information, editable artifacts, applications, people, and running processes. Actions include tool and application operations, delegation, requests for human input, and delivery.

At decision boundary $k$, the task and environment state $s_k$ evolves through agent actions and contributions from humans and ongoing processes:
\begin{equation}
  \begin{aligned}
    s_{k+1} &= F_{\mathcal E}(s_k,a_k,u_k,\epsilon_k),\\
    o_{k+1} &= O_{\mathcal E}(s_{k+1},a_k),
  \end{aligned}
  \label{eq:task-interaction}
\end{equation}
Here $a_k\in\mathcal A$ is the agent action, $u_k$ represents human instructions or direct edits, and $\epsilon_k$ captures changes from external sources and ongoing processes. The agent receives observation $o_{k+1}$ through tools, applications, or human feedback and chooses subsequent actions based on the objective and available interaction history. Humans can participate throughout the task, both by guiding decisions and by editing the same artifacts on which the agent works.

For an interaction trace $\tau$, the objective is to maximize the probability of acceptance, $\mathcal V(\tau)=1$, within budget $B$. Acceptance can assess delivered artifacts and required responses over an agreed period, using tests, external evaluation, or human judgement.

\subsection{Our View of Agency}
\label{sec:agency}
\label{sec:persistent-agency}
\label{sec:long-horizon-formulation}
\label{sec:workload-requirements}
\label{sec:principles}

We view \emph{agency} as a system's ability to choose and carry out actions that advance a user's objective, adapting its approach to observations and human guidance~\citep{sumers2024cognitive}. The challenge of exercising this ability depends on both the structure of the work and how long the system must remain responsible for it. We characterize these demands along two dimensions (Figure~\ref{fig:intro-overview}(a)).

\paragraph{Task Complexity.}
This dimension captures the number and interdependence of decisions, the branches that must be explored, and the tools and evidence that must be combined to produce an acceptable result~\citep{liu2024agentbench}. Software development, research, and design can demand substantial planning, investigation, and verification even when little time is spent waiting for external events. These demands remain even if inference and tool execution become arbitrarily fast. Complex delivery primarily challenges agency along this dimension.

\paragraph{Temporal Extent.}
This dimension captures how long the objective remains active, including periods awaiting computations, human input, or changes in the environment. A monitoring task may require few decisions while remaining relevant for days or weeks~\citep{sumers2024cognitive}. Sustained supervision therefore calls for the ability to retain the objective, recognize when new information warrants action, and respond as conditions change.

Tasks can be demanding along either or both dimensions. Persistent monitoring can combine a long elapsed horizon with relatively simple local decisions, while sustained project work can require both complex problem solving and continued responsibility. Recursive self-improvement (RSI) illustrates their intersection~\citep{zelikman2023stop}: an agent analyzes weaknesses and proposes changes, waits for training and evaluation, then uses the results to guide further work. \system{} targets tasks across these dimensions, with a common framework for complex delivery, sustained supervision, and their combination.
\section{The \system{} Agent Harness}
\label{sec:method}

\begin{figure}[!t]
\centering
\includegraphics[width=\linewidth]{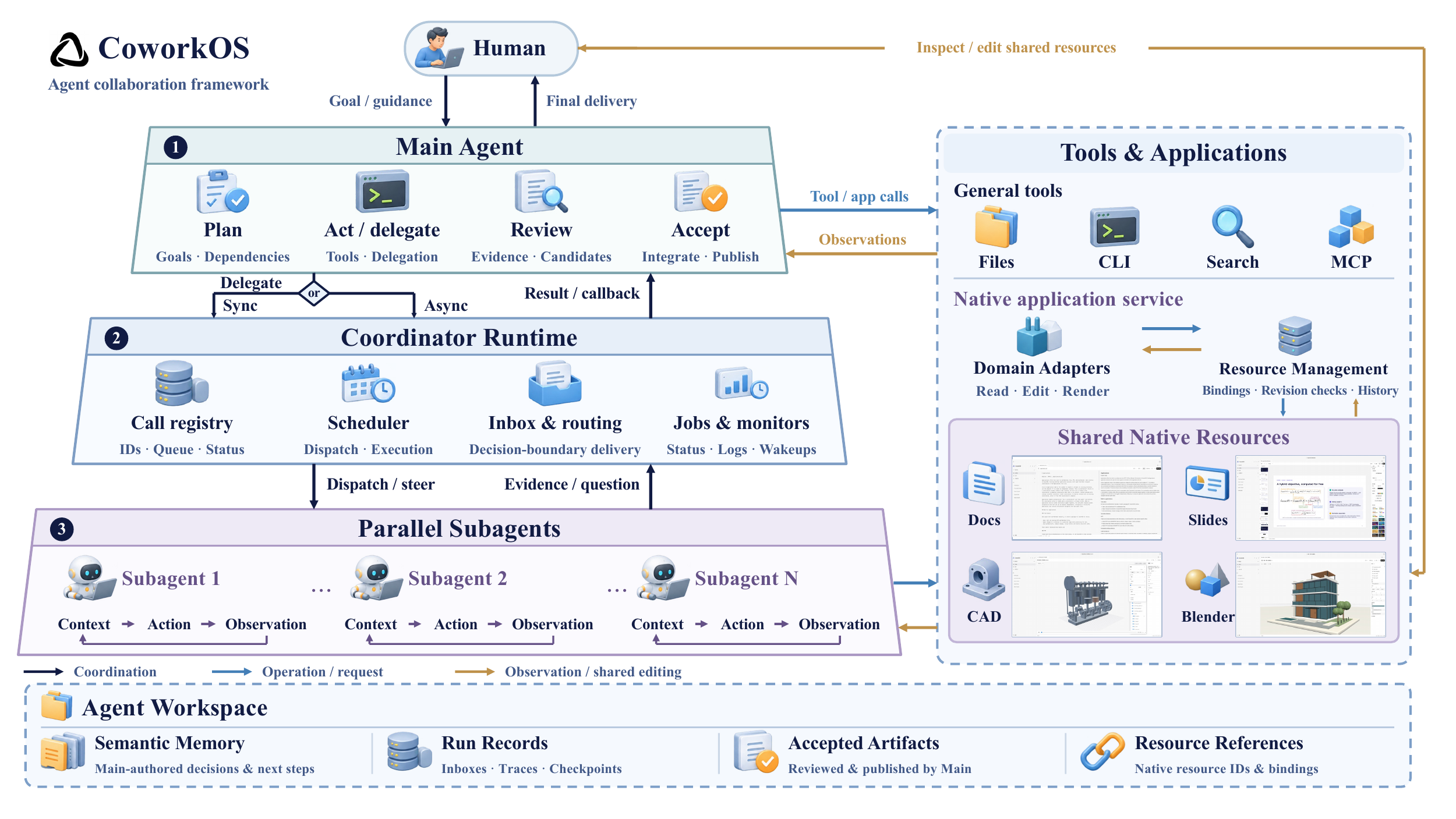}
\caption{The \system{} harness and its shared action space. Main directs and reviews work, while the Coordinator Runtime manages synchronous or asynchronous Subagent execution and feedback. Main and Subagents share tools, application services, and an agent workspace. Asynchronous delegation lets Main continue other work. Application services support native resource editing, revision management, and conflict detection.}
\label{fig:framework}
\end{figure}

In this section, we introduce \system{}, a cowork harness for long-horizon and human-in-the-loop task execution. Figure~\ref{fig:framework} shows three parts of the harness:
\begin{itemize}
  \item \textbf{Hierarchical Agent Runtime Architecture:} a three-layer runtime with communication and coordination during execution.
  \item \textbf{Scalable Application Ecosystems for Human--Agent Collaboration:} an application extension model and a shared resource protocol for human and agent contributions.
  \item \textbf{State Management for Persistent Agent Workspaces:} monitoring, checkpointing, workspace management, and context and memory mechanisms that preserve progress across executions.
\end{itemize}

\subsection{Agent Runtime Architecture}
\label{sec:coordination}
Long-horizon coworking creates a tension between global coherence and local execution. The system must preserve one accountable interpretation of the user's objective, while allowing context-heavy investigations to proceed in parallel and recover from interruption. We resolve this tension by separating authority from execution: Main owns task-level decisions, Subagents own bounded assignments, and the Coordinator makes the exchanges between them durable and observable. This organization is implemented through a hierarchical runtime (\ref{sec:hierarchical-agents}) and explicit communication paths (\ref{sec:communication}).

\subsubsection{Hierarchical Agent Organization}
\label{sec:hierarchical-agents}
To preserve a coherent global context while enabling efficient parallel and independent task execution, we adopt a hierarchical agent organization with a centralized control structure centered on Main. Main serves as the user-facing interface and the overall planner, manager, and executor, and autonomously invokes Subagents when parallel processing or independent task handling is needed. This design is motivated by three considerations: 
\begin{itemize}
  \item the full task context remains in Main, allowing it to better integrate user instructions with the interaction history
  \item context-heavy tasks that can be handled independently can be delegated to Subagents with isolated contexts, improving efficiency without interfering with Main's context
  \item concentrating most responsibilities in Main facilitates learning effective task control and coordination.
\end{itemize}

\paragraph{Main Agent.}
Main is the system's point of accountability to the user. It interprets instructions, acts directly when the task requires global context, and revises ongoing work as the user changes objectives or constraints. To support this role, the harness gives Main a comprehensive toolset and access to skill documents, with relevant capabilities and instructions loaded as needed. This allows Main to investigate, execute, and verify work within its own decision loop. At decision step $t$, it selects an action according to
\begin{equation}
  a_t^{\mathrm M}
  \sim
  \pi_{\mathrm M}
  \left(
    \cdot \mid g,h_t^{\mathrm M},\mathcal F_t
  \right),
  \label{eq:main-agent-policy}
\end{equation}
where $g$ is the user's objective, $h_t^{\mathrm M}$ is the interaction history retained in Main's context, and $\mathcal F_t$ contains newly available feedback, including execution results and user guidance.

Beyond direct execution, Main can request human input and delegate scoped work through dedicated tools. A delegation is therefore an evidence contract: Main specifies the objective, constraints, capabilities, and form of evidence expected at the boundary of the assignment. This is useful for long tool interactions whose intermediate trajectory need not enter Main's context and for independent activities that can run in parallel. Main remains responsible for interpreting the returned contribution and deciding how it affects the task.

To keep this accountability loop responsive during lengthy operations, the runtime supports synchronous and asynchronous execution. A synchronous call keeps the result on Main's calling path. An asynchronous delegation or background command returns an acknowledgement and identifier, while its outcome is durably queued in Main's Inbox and incorporated at a later decision boundary alongside user guidance. Main can continue independent work while results are pending, or wait explicitly when a result is a prerequisite for acceptance.

\paragraph{Subagents.}
Each Subagent maintains a separate working context for a scoped assignment. It can conduct extended local investigation and tool use without adding every intermediate step to Main's task-level context. For assignment $d_i$, its initial context is constructed as
\begin{equation}
  c_{i,0}
  =
  \Phi_{\mathrm S}
  \left(
    d_i,\,
    b(h_{t_i}^{\mathrm M}),\,
    \Gamma_i,\,
    \mathcal S_i
  \right),
  \label{eq:subagent-context}
\end{equation}
where $t_i$ denotes the dispatch step, $b(\cdot)$ extracts bounded recent background from Main, $\Gamma_i$ specifies the granted tools and resources, and $\mathcal S_i$ is a compact catalog of available skills. The Subagent retrieves full skill instructions and additional working materials as needed.

Within this scope, the Subagent chooses and revises a local sequence of actions as observations arrive, forming a trajectory $\tau_i$. It may explore alternatives, construct candidate outputs, and verify them with the granted capabilities. It returns a contribution $r_i=(y_i,\mathcal P_i,\sigma_i)$ consisting of a report, references to outputs and supporting records, and execution status. The report states what was done and what remains uncertain, which makes the result inspectable without importing the entire trajectory into Main.

\paragraph{Coordinator Runtime.}
The Coordinator is the control-plane boundary between these contexts. On delegation, it assembles the assignment, bounded background, and granted capabilities into a worker context. On return, it exposes the report, evidence references, and compact execution status to Main while keeping the detailed trajectory separately accessible. The Coordinator thus moves evidence across context boundaries without merging the contexts themselves.

These information exchanges can span multiple model decisions and interruptions, so their ownership and delivery state must persist beyond the current context. The Coordinator therefore maintains assignments, execution state, and pending feedback in persistent program state. Let $\xi_k$ denote this state after event $k$. Its evolution follows
\begin{equation}
  \xi_{k+1}
  =
  F_{\mathrm C}
  \left(
    \xi_k,e_{k+1}
  \right),
  \label{eq:coordinator-transition}
\end{equation}
where $e_{k+1}$ represents a delegation request, execution update, message, or terminal result, and $F_{\mathrm C}$ is the durable runtime state-transition logic. Stable identities associate guidance and returned results with the corresponding assignment and Task. Queues, leases, and acknowledgements preserve ownership and delivery state across model decisions and recoverable interruptions, so Main can continue from operational evidence rather than from an assumed in-memory history.

\subsubsection{Agent Communication and Coordination}
\label{sec:communication}

Information flows through three directional paths in the hierarchy: Main-to-Subagent, Subagent-to-Subagent, and Subagent-to-Main. We represent the communication graph as
\begin{equation}
  \mathcal G_t
  =
  \left(
    \mathcal V_t,\,
    \mathcal E_t^{\mathrm{hier}}
    \cup
    \mathcal E_t^{\mathrm{peer}}
  \right),
  \label{eq:agent-communication-network}
\end{equation}
where $\mathcal V_t$ contains Main and the Subagents participating in the Task, $\mathcal E_t^{\mathrm{hier}}$ contains interactions involving Main, and $\mathcal E_t^{\mathrm{peer}}$ contains direct Subagent--Subagent exchanges.

\paragraph{Main-to-Subagent Transmission.}
Main starts a delegated execution with a concrete objective, relevant background, constraints, expected evidence, and explicit capability grants. The Coordinator assigns a stable identity, creates the scoped worker context, and binds it to the originating Task. This handoff transfers enough context for local decisions without copying Main's complete history. During execution, Main may enqueue a focused correction, new constraint, or follow-up requirement. The worker queues guidance and incorporates it at a safe model decision boundary. Cancellation uses a separate control path that can terminate the execution.

\paragraph{Subagent-to-Subagent Transmission.}
Some assignments cannot be fully isolated: a worker may depend on an interface, assumption, or candidate held by another worker. We therefore provide a minimal peer channel within a Task and Run. A Subagent sends concise text or a file/artifact reference to a non-terminal peer through the Coordinator. The Coordinator persists it in the recipient's Inbox and returns a message identifier with queued status; at a later safe decision boundary, the recipient sees the sender identity with the message. Queueing and delivery are recorded, while Inbox state tracks consumption. Messages to terminated targets are rejected or marked undelivered. The channel is asynchronous and non-blocking, and a response is optional and expressed as another peer message. It resolves local dependencies without turning peer exchange into a second control hierarchy: Main still owns task scope, further delegation, and final acceptance.

\paragraph{Subagent-to-Main Transmission.}
When a Subagent completes or fails, it returns a report, execution status, and references to outputs, diagnostics, and supporting records. With synchronous delegation, the result remains on Main's calling path. With asynchronous delegation, the runtime first returns an acknowledgement and call identifier, allowing Main to continue independent decisions; the terminal result later enters Main's Inbox and can trigger a subsequent Main turn. Main may wait for selected assignments or for all outstanding assignments when their results are prerequisites for acceptance.

An asynchronous Subagent may request a concrete decision from Main when an ambiguity changes its approach. The request carries a bounded wait and a declared fallback assumption, so the worker can make progress without holding the runtime indefinitely. Completed contributions, decision requests, user messages, and relevant background events enter Main through the same task-level integration path. Subagent contexts do not receive user inputs directly; Main assesses their effect on the global objective and decides whether to revise, continue, or cancel delegated work.

\paragraph{Message Delivery and Shared Evidence.}
Across the three paths, a message carries the smallest useful unit of coordination and points to evidence that can be inspected separately. The Task workspace exposes selected references and accepted artifacts without granting workers unrestricted write access (Figure~\ref{fig:workspace}). The Coordinator associates each message with its Task or assignment, delivers it at a model decision boundary, and preserves deferred messages for recovery. Agents therefore exchange focused information while retaining a path to the evidence needed for local execution or task-level integration.

Synchronous results stay on the calling path. Asynchronous work returns an acknowledgement and identifier before its terminal result enters the recipient's Inbox. The durable work queue records assignment and execution state; leases and acknowledgements allow pending feedback to be recovered without duplicating callbacks or contributions. These mechanisms let focused contexts progress independently while preserving one task-level interpretation in Main.

\subsection{Scalable Application Ecosystems for Human--Agent Collaboration}
\label{sec:applications}
\label{sec:tools}

Long-horizon coworking rarely stays inside one application. The same artifact may be discussed in conversation, edited by a user, inspected by one agent, and refined by another. The central integration problem is therefore continuity of resource state: each participant must act on a known version of the artifact, and a contribution must remain attributable when the artifact changes underneath it. We build an application ecosystem inside \system{} around this invariant.

\begin{figure}[!t]
\centering
\includegraphics[width=\linewidth]{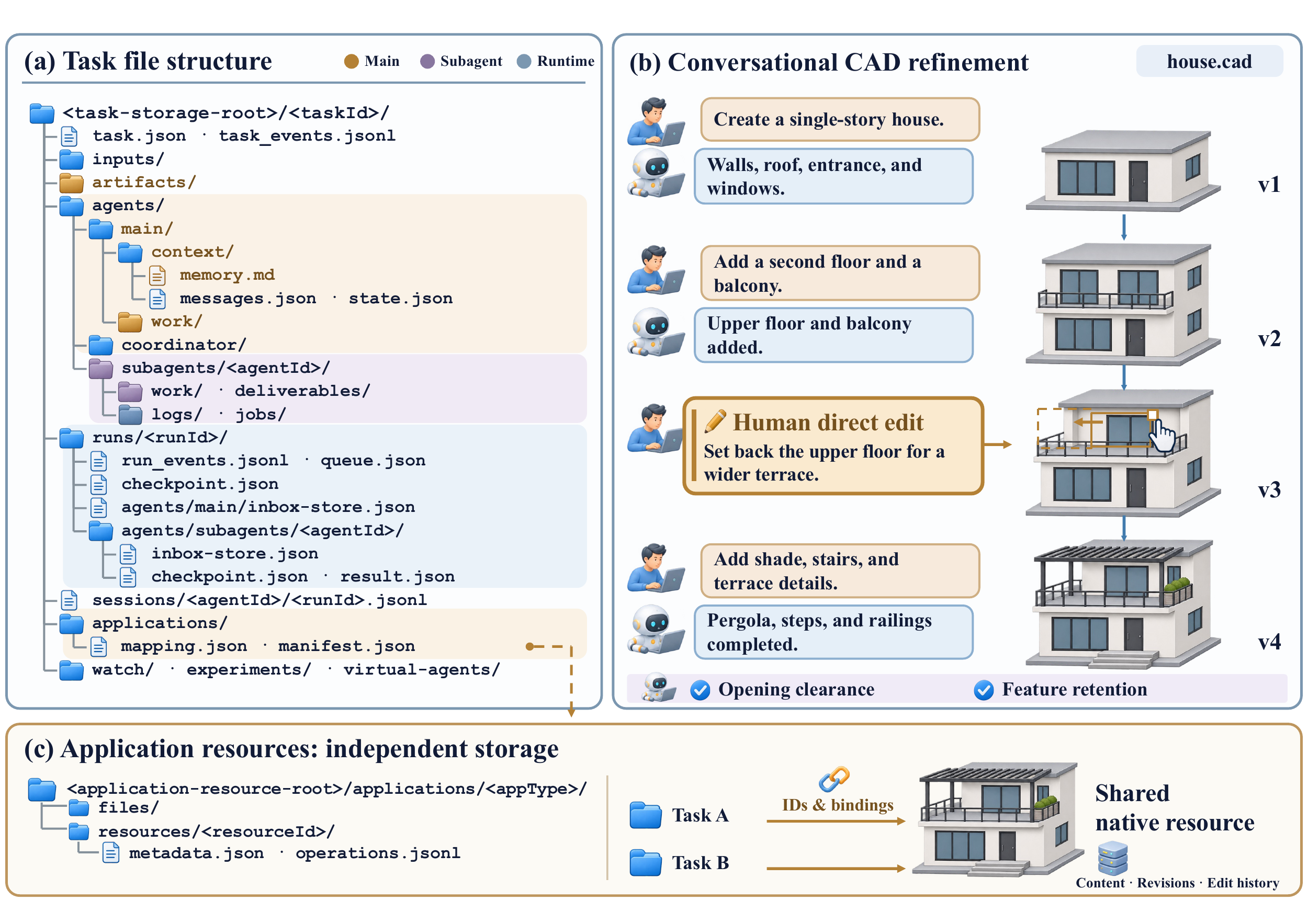}
\caption{Task workspace structure and collaborative editing. \textbf{(a)} Selected file paths and their owners or record managers across agent, Run, and session scopes. \textbf{(b)} Iterative CAD refinement through user instructions, agent updates, and direct human edits to a shared model. \textbf{(c)} Task references and bindings link to independently stored application resources with persistent content, revisions, and edit history.}
\label{fig:workspace}
\end{figure}

To make this ecosystem extensible, we first define a common resource protocol for application integration. It gives the runtime a shared way to identify resources, apply and validate edits, and preserve revisions as work moves between users, agents, and successive executions. On top of this protocol, we support two integration paths: native applications managed by \system{} and existing software connected through MCP. The two paths differ in how capabilities enter the workspace while sharing the same Task and resource lifecycle. Figure~\ref{fig:workspace} shows collaboration through a native application resource.

\subsubsection{Resource Protocol}
\label{sec:resource-protocol}

For a platform-managed resource, identity and revision are first-class because an edit without a base state cannot be evaluated after concurrent work. We represent it as $r=(\iota,\kappa,x,v)$, where $\iota$ is its identity, $\kappa$ its kind, $x$ its application-owned content, and $v$ its committed revision. The resource has a lifecycle independent of any one conversation or application window; Tasks and successive Runs hold references to the same evolving artifact.

The platform manages identity, references, revisions, and history, while each adapter preserves the semantics of its domain. Text edits and CAD parameter updates can therefore remain different operations while following one admission rule. For a new operation $u$ that passes domain validation,
\begin{equation}
  \operatorname{Commit}(r,u,\hat v,\alpha)
  =
  \begin{cases}
    (\iota,\kappa,T_{\kappa}(x,u,\alpha),v+1), & \hat v=v,\\
    \text{conflict}, & \hat v\ne v,
  \end{cases}
  \label{eq:resource-commit}
\end{equation}
where $\hat v$ is the operation's base revision, $\alpha$ identifies the human or agent actor, and $T_{\kappa}$ applies the operation according to the adapter's semantics. Both human and agent edits follow this rule. An accepted change advances the revision and records its actor and summary. A stale proposal returns a conflict rather than silently overwriting intervening work; the proposal remains available for reconsideration against the current state. Revision control is thus the mechanism that turns mixed human--agent editing into an auditable sequence of accepted changes.

Main can grant a Subagent permission to read a resource or prepare a private candidate. The candidate remains separate from the platform-managed artifact while the Subagent works, so exploration does not become an unreviewed shared edit. The Subagent returns the candidate with a report and supporting evidence; Main decides whether it should enter the artifact. The protocol then rechecks the candidate against Equation~\ref{eq:resource-commit}. If the baseline has changed, the candidate is preserved for reconsideration instead of being applied implicitly.

\subsubsection{Native Applications}
\label{sec:native-applications}

Native applications keep the resource under the platform's control, which lets direct human editing participate in the same coordination loop as agent actions. A user can revise a paragraph, move an object, or change a parameter to express a preference; the agent then observes that committed state through domain tools. Conversely, an accepted agent change updates the editor. Human edits are therefore task guidance expressed through the artifact itself, rather than a separate instruction channel.

The agent's application context includes the current resource and changes it has yet to acknowledge. For agent $a$, we write
\begin{equation}
  c_{t,\mathrm{app}}^{a}
  =
  \Phi_{\kappa}
  \left(
    r_t,s_t,\Delta_{v_a:v_t}^{\neg a}
  \right),
  \label{eq:application-context}
\end{equation}
where $v_t$ is the current revision, $v_a$ is the agent's last acknowledged revision, $s_t$ is the saved selection, and $\Delta_{v_a:v_t}^{\neg a}$ contains intervening changes by other actors. The application supplies a bounded summary through $\Phi_{\kappa}$; the agent retrieves detailed content through domain tools when needed. Change summaries exclude the agent's own edits, which are already represented in its interaction history. Human contributions can therefore enter the working context without loading the full application state at every decision.

Explicit application binding keeps the agent's scope focused as the workspace gains integrations. A binding associates the current resource with its tools, skills, and workflow instructions; when the binding changes, the runtime updates this scope and deactivates the previous application's tools. This prevents capabilities from unrelated applications from silently entering a decision context.

\begin{figure}[!htbp]
\centering
\includegraphics[width=\linewidth]{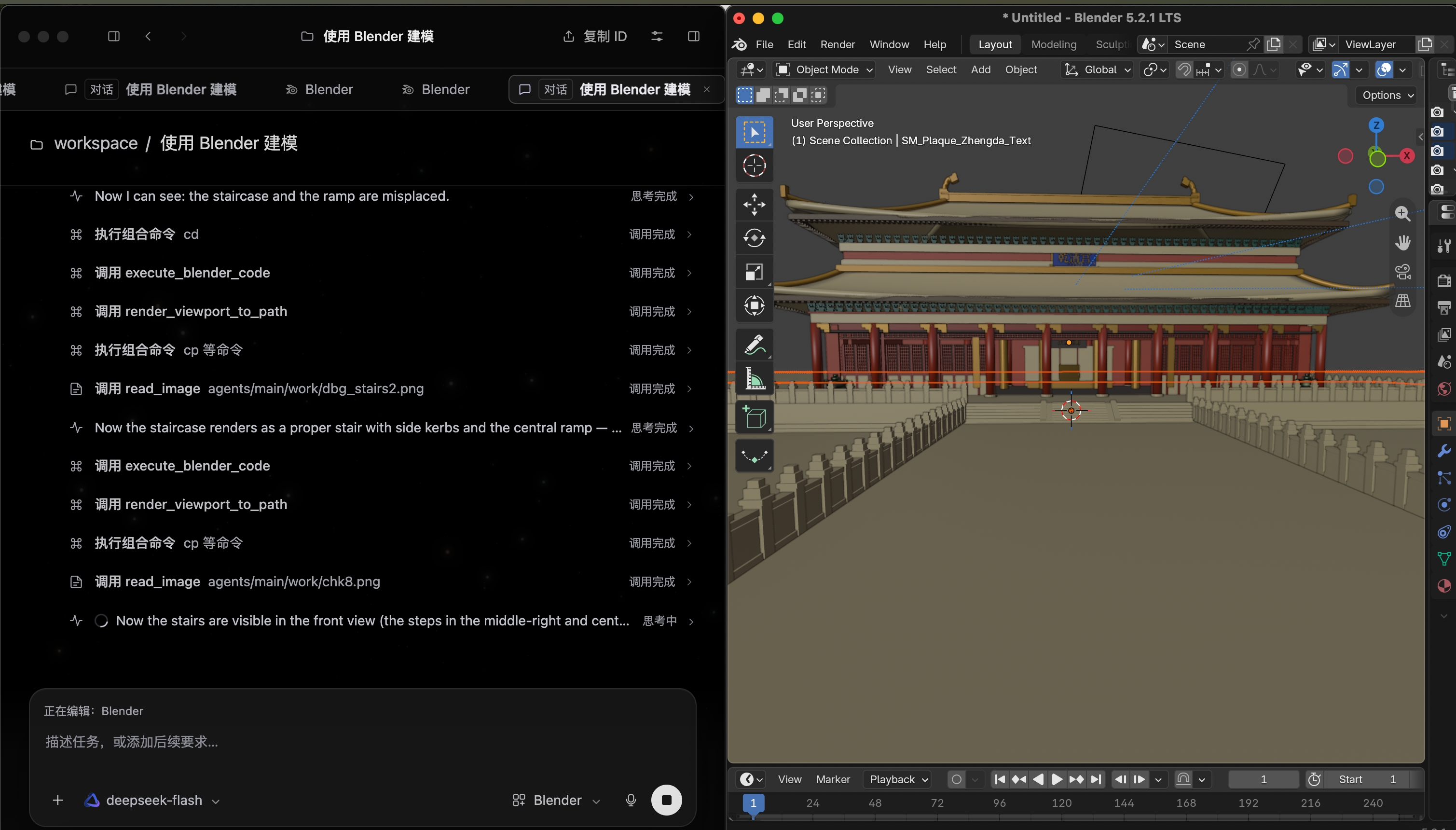}
\caption{Blender integration in \system{}. The agent conversation (left) records code execution, updates to workspace files, and image inspection. Blender's native viewport (right) shows the resulting palace model in an editable 3D scene.}
\label{fig:blender-workbench}
\end{figure}

Domain operations address existing objects and parameters and return feedback on their effects. In Blender, stable object identities support targeted changes, and Python execution retains the existing scene by default. An agent can construct a scene, a user can adjust and save an object's transform or material, and the agent can read the updated scene before refining other objects (Figure~\ref{fig:blender-workbench}). In CAD, editable source and parameter values preserve the design; geometry diagnostics and rendered views give both participants evidence for evaluating further changes.

Our native applications cover text documents, presentations, parametric CAD, and Blender projects. Each adapter supplies the editing and inspection operations for its domain. Import and export connect these editable resources to ordinary files and Task deliverables.

\subsubsection{MCP-Integrated Software}
\label{sec:mcp-software}

MCP integration preserves an existing application's ownership and interface while making its capabilities available inside the Task. A plugin contributes the software's tools, domain skills, and workflow instructions, together with an adapter and interface integration that connect it to the runtime. The application can evolve independently while the Task still records which session and capabilities were used.

Selecting a software session binds its capabilities to the agent. The application owns the MCP service lifecycle, so the session can persist across individual agent executions, and users retain access to its interface throughout the work. Our integrations use external native windows for FreeCAD and Blender and embedded native views for CAD and OfficeHarness alongside the ongoing Task.

Document ownership depends on the integration. An MCP service that returns declared resource operations can participate in the shared protocol, including revision checks and history. An integration that directly controls external software instead binds a session descriptor; the application retains its own document, saving, and concurrency semantics. This boundary keeps the platform from claiming guarantees that an external application does not expose.

Browser and desktop interaction are an orthogonal interaction mode for workflows that require graphical interfaces, rather than a third resource protocol. Agents use semantic or visual observations to operate Task-owned browser pages or a selected desktop window, then inspect subsequent observations or resulting artifacts. Users can pause desktop control and take over the window, moving between automated operations and direct interaction within the same Task.

\FloatBarrier
\subsection{State Management for Persistent Agent Workspaces}
\label{sec:continuity}

A long-horizon Task must retain progress across executions and supply a focused context for each decision~\citep{sumers2024cognitive,shinn2023reflexion}. Building on the task formulation in Section~\ref{sec:formulation}, we preserve ongoing work and deferred feedback in runtime state, keep artifacts and evidence in the workspace, and reconstruct the agent's context from memory and current task state.

We represent retained Task state as $z_k=(W_k,P_k,R_k)$, where $W_k$ contains working objects and artifact references, $P_k$ records progress, decisions, and evidence, and $R_k$ records ongoing executions, waiting conditions, and pending feedback. A Run records one concrete execution within the Task.

\subsubsection{Persistent Execution}
\label{sec:persistent-execution}

The runtime stores $R_k$ independently of the model process, keeping ongoing work, waiting conditions, and pending feedback with the Task across turns and restarts. Computations and observation sources can run while Main is idle. The runtime resumes Main when a result, external event, or scheduled review calls for another decision.

\paragraph{Background Execution and Monitoring.}
Long computations run as background Jobs with persistent identities, process records, and logs. A training run or build can continue on the running host across conversation turns, context resets, and Main restarts. Main inspects process liveness, elapsed time, log activity, and recent output to decide whether to wait or intervene. If a process disappears without a recorded exit, its outcome remains uncertain; its logs remain available for investigation.

Deterministic source scripts monitor external state by comparing samples with a saved baseline. Unchanged observations require no model invocation. Candidate events go directly to Main or undergo semantic analysis first. Sampling proceeds independently of model analysis, and source cadence is separate from Main's scheduled goal reviews, allowing frequent observation with few idle model calls.

Before parking, Main records its progress, waiting conditions, and intended next action. A continuing goal must have an active observation source or an explicit review time; the runtime rejects a wait that provides neither. In-session goals retain the current Task, while independent background goals have their own Task and Main. Local services run the observation sources and persist source and goal state for continuation after a restart.

\paragraph{Checkpointing and Recovery.}
A source commits its updated cursor together with emitted events and retains those events until a durable Inbox accepts them. Stable identities suppress duplicate delivery during replay. The runtime acknowledges feedback after it enters Main's saved context (Section~\ref{sec:communication}), so observations that Main has not yet processed remain available after a restart.

At completed turns, the runtime checkpoints the agent's message trajectory, compaction summary, and the position separating summarized and retained messages. An unfinished Run can resume from this state after a recoverable interruption; Main's session history and context snapshots also support task-level recovery. A supervisor reactivates an absent Main when eligible work is pending. It coordinates execution ownership and respects explicit pauses and waits for human input. External work may have progressed since the checkpoint, so the resumed agent can inspect current process, artifact, or application state before deciding whether to repeat an operation.

The runtime checks for active Jobs and delegations, undelivered feedback, unanswered requests, and recorded wake conditions, together with a completed assistant response, when assessing Task completion. Main evaluates whether the result satisfies the objective. A finished response can therefore leave the Task waiting for work that still requires attention.

\subsubsection{Workspace and Artifact Management}
\label{sec:workspace}

The workspace retains working objects $W_k$ alongside the progress records and supporting evidence in $P_k$. Agent working materials, Run records, and published artifacts occupy separate scopes under the Task (Figure~\ref{fig:workspace}). Agent working areas persist across Runs, so later executions can reuse existing files. Each Run retains its own events, checkpoints, and results for inspection of earlier work and interrupted attempts.

Agents can inspect files, contributions, and execution records throughout the Task. By default, they write only to their own working area; explicit capabilities can extend that scope to a project being edited. An agent can read another agent's findings while developing its own drafts and intermediate outputs, retrieving detailed evidence through shared references as needed.

Subagents place reusable outputs in their delivery areas and return references with conclusions and supporting evidence. Main reviews these candidates, integrates contributions as needed, and publishes selected results to the Task artifact area. Later assignments can start from reviewed artifacts and inspect the original candidate files or their producing Run records when needed.

Native application resources have their own storage, revisions, and lifecycle. The Task keeps references to them, and contributions follow the review and update protocol in Section~\ref{sec:applications}. Both files and application resources remain available when the agents and contexts working on them change.

\subsubsection{Memory and Context Management}
\label{sec:memory}

The agent uses persistent Task state and available history $h_k$ within a bounded decision context:
\begin{equation}
  c_k=\Phi(g,h_k,z_k), \qquad \operatorname{tokens}(c_k)\le L,
  \label{eq:general-agent-context}
\end{equation}
where $\Phi$ selects and assembles information for the current decision, and $L$ is the context capacity. To reconstruct Main's context as the history grows, we combine the objective and consequential decisions in durable memory with current runtime and goal state. Detailed evidence remains accessible in the workspace for decisions that require more than the reconstructed context contains.

\paragraph{Task Memory.}
Main maintains a memory document with the objective, completed work, next actions, and important information or constraints. The document preserves decisions needed in later phases. A compact runtime view supplies execution status, pending feedback, and waiting conditions, allowing Main to reconcile its remembered plan with work that is still active or has since returned. Optional user preferences and project experience enter through scoped projections with a context budget. Project skills are listed compactly and loaded in full as needed.

\paragraph{Context Reconstruction and Compaction.}
At phase boundaries, Main can refresh its context for the work ahead. Before requesting a reset, it is instructed to update its memory with the progress and decisions needed for continuation. Once the current tool batch settles, the runtime applies the reset and constructs the task context as
\begin{equation}
  c_{t,\mathrm{task}}^{\mathrm M}
  =
  \Phi_{\mathrm{ctx}}
  \left(
    m_T,\rho_T(t),d_T(t)
  \right)
  \oplus h_{t,\mathrm{recent}}^{\mathrm M},
  \label{eq:task-context-reconstruction}
\end{equation}
where $m_T$ is Main's semantic memory, $\rho_T(t)$ projects the runtime and structured goal state, $d_T(t)$ summarizes recorded experiments when available, and $\oplus$ concatenates retained recent history. An explicit reset starts from the reconstruction summary with $h_{t,\mathrm{recent}}^{\mathrm M}=\varnothing$; automatic compaction retains a recent segment at a safe turn boundary. When the summary is insufficient, Main can inspect workspace materials and full interaction logs. It can also read current application state to account for changes made during a pause.

Within an assignment, the runtime compacts a Subagent's execution trajectory as the estimated context approaches the model's capacity. It summarizes an older prefix while retaining recent interactions and effective system instructions. Before adopting the compacted state, it persists the summary and its message cursor together so both can be restored consistently with the execution checkpoint. If summarization fails, the previous history remains available.

\paragraph{Experimental History.}
A Task-level ledger records experiment objectives, delegated Runs, candidate variants, artifacts, and metrics with source references. It distinguishes measurements read from run-produced metric files from values stated only in narrative reports. Once multiple measured rounds are available, a compact digest of initial and recent results accompanies context reconstruction and callbacks for the experiment. Main can compare these results, inspect the associated artifacts, and select the next experiment after a context reset. Section~\ref{sec:rsi} evaluates this recursive self-improvement loop.

\section{Experiments}
\label{sec:experiments}

We evaluate \system{} through standardized benchmarks, real-world application cases, and iterative improvement experiments. Section~\ref{sec:benchmarks} examines task performance across four benchmark domains; Section~\ref{sec:cases} considers the completeness, consistency, and usability of delivered artifacts; Section~\ref{sec:rsi} studies aggregate progress and task-level changes across improvement rounds. 

\subsection{Benchmark Evaluation}
\label{sec:benchmarks}

\subsubsection{Evaluation Setup}
\paragraph{Benchmarks.}
To evaluate complex, long-horizon, end-to-end productivity work, we select twelve benchmarks covering four capabilities. \textbf{Tool Use} includes SkillsBench~\citep{skillsbench2026}, Tau-3 Bench Banking~\citep{tauknowledge2026}, and AutomationBench~\citep{automationbench2026}, evaluating tasks carried out through skills, tools, and interactive operations. \textbf{Productivity} includes JobBench~\citep{jobbench2026}, Harvey Legal Agent Bench~\citep{harveylab2026}, GDPVal~\citep{gdpval2025}, OfficeQA Pro~\citep{officeqa2026}, and WorkspaceBench-Lite~\citep{workspacebench2026}, covering business work, professional materials, and office tasks. \textbf{Search} includes WideSearch~\citep{widesearch2025} and BrowseComp~\citep{browsecomp2025}, assessing information retrieval and synthesis. \textbf{ML Coding} includes MLE-Bench-Lite~\citep{chan2025mlebench} and MLS-Bench~\citep{mlsbench2026}, assessing machine-learning implementation and experiment execution. Together, these domains cover complementary stages of practical work, from acting on an environment and processing information to delivering a result.

\paragraph{Model Configuration.}
We evaluate GLM-5.3 with High thinking effort, a 256K-token context window, and a maximum output length of 64K tokens per response.

\paragraph{Harnesses.}
We compare \system{} with three mainstream open-source harnesses---OpenHands~\citep{wang2025openhands}, Codex, and DeepSeek Harness (DSH)---under the above task and model configurations, reporting each benchmark's evaluation metric.

Benchmark-specific task scopes, scoring procedures, and execution limits are summarized in Appendix~\ref{sec:benchmark-protocols}. The scores describe the actual configurations used in these runs.

\subsubsection{Evaluation Results}
Table~\ref{tab:benchmarks} summarizes results across the four domains. \system{} ranks among the top two harnesses in every domain average, achieving the highest averages in Productivity and ML Coding while closely matching DSH in Tool Use and OpenHands in Search. It also achieves the best results on SkillsBench, JobBench, WideSearch, and both ML Coding benchmarks, demonstrating broad competitiveness across tool execution, knowledge work, information retrieval, and machine-learning experiments. Taken together, these results give \system{} the strongest overall performance among the four evaluated harnesses and provide empirical support for its ability to handle complex, end-to-end productivity tasks.

\begin{table}[!t]
\centering
\small
\renewcommand{\arraystretch}{1.18}
\captionsetup{skip=4pt}
\setlength{\tabcolsep}{6pt}
\begin{tabularx}{\linewidth}{@{}lYccc|c}
\toprule
\textbf{Domain} & \textbf{Benchmark} & \textbf{OpenHands} & \textbf{Codex} & \textbf{DeepSeek Harness} & \textbf{\system{}} \\
\midrule
\multirow{4}{*}{\textit{Tool Use}} & SkillsBench & \underline{63.30} & 53.48 & 56.75 & \textbf{64.34} \\
 & $\tau^3$-Bench Banking & 40.20 & 46.39 & \textbf{55.67} & \underline{49.48} \\
 & AutomationBench & 50.50 & 30.33 & \textbf{53.50} & \underline{51.83} \\
 & \cellcolor{atrialight}Average & \cellcolor{atrialight}51.33 & \cellcolor{atrialight}43.40 & \cellcolor{atrialight}\textbf{55.31} & \cellcolor{atrialight}\underline{55.22} \\
\midrule
\multirow{6}{*}{\textit{Productivity}} & JobBench & 55.02 & 23.42 & \underline{56.15} & \textbf{59.88} \\
 & Harvey Legal Agent Bench & \textbf{23.75} & 12.70 & 22.30 & \underline{22.74} \\
 & GDPVal (Elo) & 1122 & 925 & \underline{1157} & \textbf{1164} \\
 & OfficeQA Pro & \underline{66.20} & \textbf{66.30} & 62.30 & 65.90 \\
 & WorkspaceBench-Lite & 64.54 & 58.81 & \textbf{68.76} & \underline{67.14} \\
 & \cellcolor{atrialight}Average & \cellcolor{atrialight}55.27 & \cellcolor{atrialight}40.11 & \cellcolor{atrialight}\underline{56.13} & \cellcolor{atrialight}\textbf{57.54} \\
\midrule
\multirow{3}{*}{\textit{Search}} & WideSearch & 81.43 & \underline{82.05} & 72.48 & \textbf{82.47} \\
 & BrowseComp & \textbf{83.00} & 75.50 & 80.80 & \underline{81.80} \\
 & \cellcolor{atrialight}Average & \cellcolor{atrialight}\textbf{82.22} & \cellcolor{atrialight}78.78 & \cellcolor{atrialight}76.64 & \cellcolor{atrialight}\underline{82.14} \\
\midrule
\multirow{3}{*}{\textit{ML Coding}} & MLE-Bench-Lite & 60.40 & 67.28 & \underline{76.70} & \textbf{76.90} \\
 & MLS-Bench & 28.91 & 29.05 & \underline{35.57} & \textbf{35.62} \\
 & \cellcolor{atrialight}Average & \cellcolor{atrialight}44.66 & \cellcolor{atrialight}48.17 & \cellcolor{atrialight}\underline{56.14} & \cellcolor{atrialight}\textbf{56.26} \\
\bottomrule
\end{tabularx}
\caption{Benchmark evaluation across twelve tasks in four domains. Each domain ends with an equally weighted Average over the listed benchmarks. GDPVal reports Elo over all 220 scheduled tasks; its tie-adjusted win rate (percent) enters the Productivity average (Appendix~\ref{sec:benchmark-protocols}). Bold and underlining mark the highest and second-highest scores per row, including ties.}
\label{tab:benchmarks}
\end{table}

\paragraph{Tool Use.}
\system{} demonstrates strong performance across Tool Use, with a domain average of 55.22 that closely matches DSH's 55.31 and exceeds those of OpenHands and Codex. It ranks first on SkillsBench with a score of 64.34, highlighting its ability to complete tasks through skills. On Tau-3 Bench Banking and AutomationBench, it joins DSH in the top two, maintaining competitive performance in interactive tool use and automation. Together, these strengths span skill use, banking interactions, and automated execution, supporting the tool operations required by complex workflows across different task settings.

\paragraph{Productivity.}
For practical knowledge work, \system{} achieves the highest Productivity average of 57.54, ahead of DSH's 56.13, demonstrating broad strength across professional tasks. It leads all evaluated harnesses on JobBench with a score of 59.88 and ranks among the top two on Harvey, WorkspaceBench-Lite, and GDPVal, covering business, legal, and office scenarios. Its OfficeQA Pro performance is also close to that of OpenHands and Codex. This combination of results shows broad support for information processing and task delivery across diverse work settings, aligning with Cowork's goal of addressing practical productivity needs.

\paragraph{Search.}
Strong search performance is evident in both individual results and consistency across tasks. \system{} achieves the best result on WideSearch with a score of 82.47 and ranks among the top two on BrowseComp, outperforming Codex and DSH on both benchmarks while its domain average of 82.14 closely matches OpenHands' 82.22. These results demonstrate strong information acquisition that complements tool execution and professional work. For Cowork tasks requiring external information, search supplies evidence and context for subsequent analysis and artifact creation, connecting information retrieval, content integration, and final delivery within a broader workflow.

\paragraph{ML Coding.}
On both MLE-Bench-Lite and MLS-Bench, \system{} achieves the best results among the evaluated harnesses with scores of 76.90 and 35.62, respectively, demonstrating strong machine-learning programming and experiment execution capabilities. It joins DSH in the leading group on both benchmarks, with both harnesses outperforming OpenHands and Codex, showing a consistent advantage across the evaluated tasks. These results extend the capability coverage from general-purpose tools and knowledge work to specialized technical workflows. They also show that \system{} remains competitive in tasks requiring implementation and experimentation, supporting Cowork's ability to undertake end-to-end machine-learning work.

\FloatBarrier
\subsection{Qualitative Analyses on Real-World Tasks}
\label{sec:cases}

\begin{figure}[p]
\centering
\setlength{\abovecaptionskip}{6pt}
\includegraphics[width=\linewidth,height=0.94\textheight,keepaspectratio]{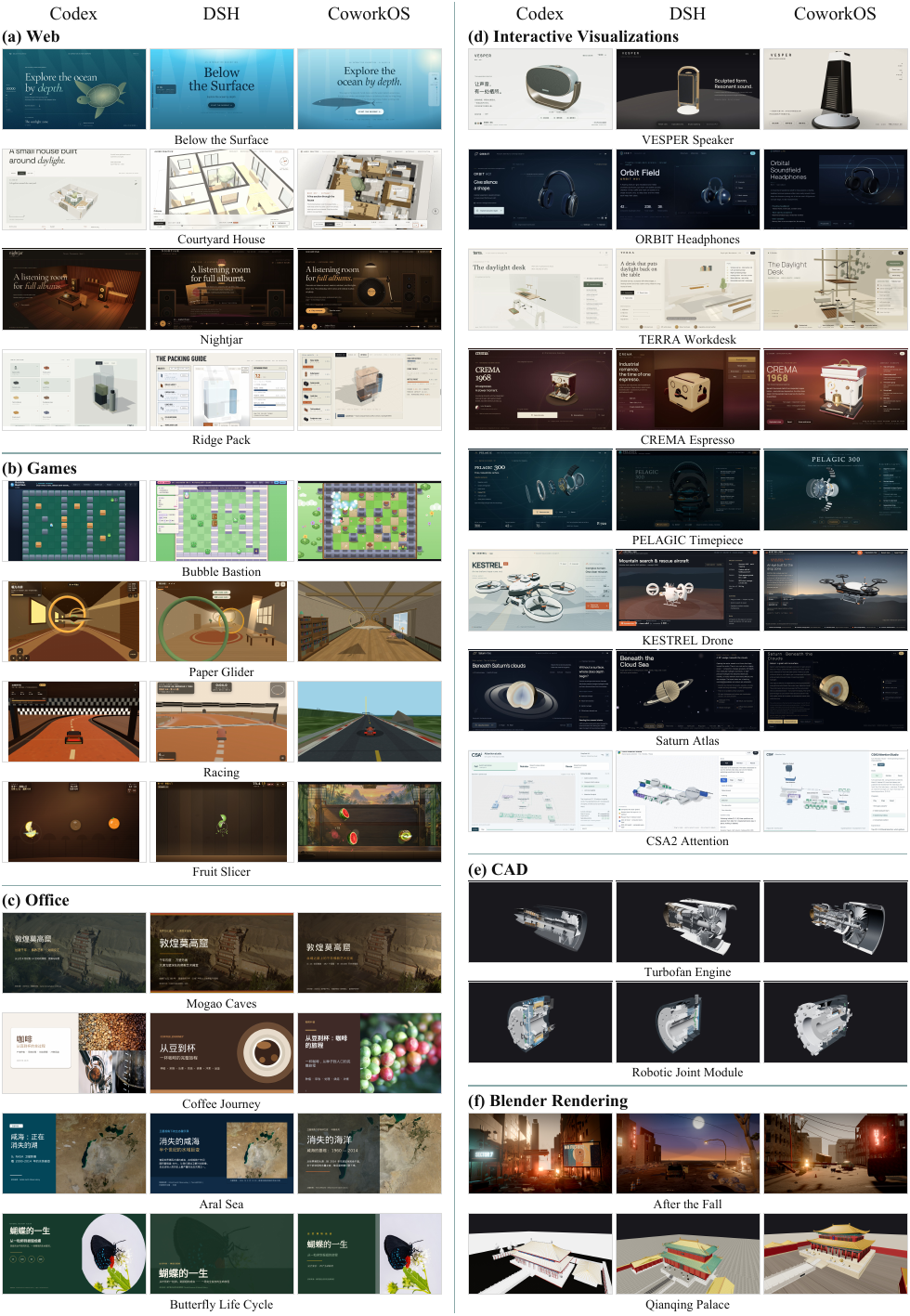}
\caption{Qualitative comparisons across 24 cases: (a) frontend websites, (b) games, (c) Office presentations, (d) interactive visualizations, (e) CAD, and (f) Blender rendering. Each case shows Codex, DeepSeek Harness (DSH), and \system{} from left to right. Complete interactive artifacts are available in the \href{https://atria-harness-gallery.chengjunpan505.chatgpt.site/\#gallery}{online showcase}.}
\label{fig:four-category-comparison}
\end{figure}

We examine how \system{} turns task requirements into visually organized, interactive deliverables. Figure~\ref{fig:four-category-comparison} covers 24 cases: four frontend websites, four games, four Office presentation tasks, eight interactive visualizations, two CAD assemblies, and two Blender scenes. The following category analyses are grounded in representative observations from inspecting the final deliverables. Content-page comparisons, inspection coverage, and additional interaction evidence are documented in Appendix~\ref{sec:case-protocols}.

\subsubsection{Frontend Websites}
{\bfseries The courtyard-inspection and packing-feedback cases show how \system{} consolidates object information and gives a more specific explanation when an operation is blocked (Figure~\ref{fig:four-category-comparison}a; interaction details in Figure~\ref{fig:web-interaction-details}).} In \emph{Courtyard House}, all three implementations support Kitchen inspection and retain the viewing settings. Codex offers an eye-level courtyard view with settings in the bottom controls, while DSH labels the room area and notes that settings are preserved. \system{} brings area, dimensions, time, section height, and material scheme together in the room card, making them directly available during inspection. In \emph{Ridge Pack}, Codex reports that the picnic blanket cannot fit and suggests removing a larger item or using a larger pack. The shown DSH view lists compartment coordinates for packed items, with the blanket still unpacked. \system{}'s rejection message additionally explains why spare aggregate volume is insufficient: the space is fragmented, and no placement clears the neighboring objects. It then suggests reordering the items or removing a bulky neighbor.

\subsubsection{Games}
{\bfseries \system{} combines more detailed scene construction with readable gameplay feedback (Figures~\ref{fig:four-category-comparison}b and~\ref{fig:game-feedback-details}).} In \emph{Paper Glider}, Codex emphasizes luminous rings and a corridor, and DSH adds a desk, lamp, and rug; \system{} organizes bookcases, curtains, seating, ceiling beams, and shadows into a recognizable study. Its result dialog also explains the missed-ring allowance and provides a working retry action. In \emph{Fruit Slicer}, all three show separated fruit and score feedback. Codex and DSH use simplified fruit against wooden backgrounds, while \system{} integrates detailed fruit interiors, blade trails, and juice effects with a bamboo-framed mountain dojo; normal pointer input visibly advances both the cut count and score.

\subsubsection{Office}
{\bfseries \system{} presents information more clearly through metric grouping and relation diagrams in the inspected Office pages.} In \emph{Mogao Caves}, Codex's descriptions extend below the metric-card borders, while DSH's quantity and duration suffixes wrap into the explanatory text. \system{} keeps each value, label, and explanation inside its card, making their association easier to follow. In \emph{Aral Sea}, Codex's relation diagram places reduced inflow and evaporation sequentially in one chain, while DSH explains the causes in three text cards. \system{} instead draws them as separate branches converging on lake shrinkage, allowing readers to trace the two displayed paths directly. These observations support clearer grouping and connections in the shown pages. Content-page comparisons and page-level observations appear in Appendix~\ref{sec:case-protocols} (Figure~\ref{fig:office-content-details}).

\subsubsection{Interactive Visualizations}
{\bfseries In structural inspection and mechanism explanation, \system{} stands out in expanded-view framing and state legibility (Figure~\ref{fig:four-category-comparison}d).} In \emph{CREMA Espresso}, expansion moves the camera back and retracts the long introduction, separating the lid, boiler group, brew head, and base vertically while keeping component descriptions visible. Codex also provides a clear exploded assembly and eight descriptions; DSH requires manual zoom-out to fit the shell, which still partly occludes the boilers. The contrast is more direct in \emph{CSA2 Attention}: in Reuse mode, \system{} presents selected positions in large front-facing labels, alongside yellow Main KV, coral Top-K, and an inactive indexing branch. Codex likewise shows cache preservation and inactive indexing, but its oblique position labels are smaller; DSH's low-angle layout crowds labels around Main Q, SWA KV, and Selection. These choices make geometry, provenance colors, and stage explanations easier to cross-reference within the displayed view. Operation details and case-specific limits appear in Appendix~\ref{sec:case-protocols} (Figure~\ref{fig:interactive-inspection-details}).

\subsubsection{CAD}
{\bfseries In the CAD cases, \system{} extends the core structure to present richer assembly detail (Figure~\ref{fig:four-category-comparison}e; detail views in Figure~\ref{fig:cad-inspection-details}).} In \emph{Turbofan Engine}, Codex emphasizes the shaft, blade groups, and exhaust within a transparent casing. DSH already has detailed multistage blade rows in section, while its exterior mainly exposes annular bands and rods. By comparison, \system{} places bent tubes, annular components, and block-like accessories around the blade rows, making both the core and peripheral structure visible together. In \emph{Robotic Joint Module}, its longitudinal ribs, populated flange fasteners, and side connection detail provide more explicit external assembly interfaces than DSH's smooth housing and flange holes; Codex likewise includes ribs, flange fasteners, and a layered interior.

\subsubsection{Blender Rendering}
{\bfseries Material distinctions and spatial depth are prominent strengths of \system{}'s Blender artifacts in these cases (Figure~\ref{fig:four-category-comparison}f).} In \emph{Qianqing Palace}, Codex includes layered roofs, window divisions, and balustrades, but broad roof and wall surfaces appear pale. DSH provides colored tiles and painted eaves, with much of the main facade formed by continuous red door and window panels. \system{} distinguishes golden roofs, freestanding red columns, lattice panels, blue--green eave decoration, and pale stonework, with a visible separation between the colonnade and wall. In \emph{After the Fall}, damaged multi-storey buildings enclose a street connecting foreground vehicles, a mid-street bus, and distant buildings; overhead cables and a hotel sign link the facades. Codex likewise creates a richly detailed urban corridor, using foreground signage, rubble, and street equipment to strengthen enclosure. DSH concentrates a rusty vehicle, low-rise ruins, and lighting around a more open shelter setting. These observations support distinguishable architectural components and connected environmental composition. Additional views and inspection limits appear in Appendix~\ref{sec:case-protocols} (Figure~\ref{fig:blender-scene-details}).

\subsection{AI-Automated Post-Training}
\label{sec:rsi}

\subsubsection{Motivation}
Recursive self-improvement (RSI) offers an important direction for the continued development of AI capabilities: systems actively participate in optimizing their own capabilities and improvement mechanisms~\citep{zelikman2023stop}. Improved systems can then drive subsequent improvements, potentially forming a reinforcing feedback loop ~\citep{duan2026rsi}. AI-automated post-training is a representative practical direction within RSI, in which AI agents undertake the research and execution of model post-training to continually improve model capabilities.

This task brings complex delivery and sustained supervision together around a single model-improvement objective, exemplifying the sustained project work in Figure~\ref{fig:intro-overview}(a). Failure analysis, pipeline modification, and training decisions are interdependent and evolve as new evidence emerges; delayed training feedback and progress across rounds require continued maintenance of code, data lineage, and execution state. Both challenges closely match \system{}'s core design: model-directed execution supports flexible research decisions, while continuity across contexts and time sustains long-running work. Automated post-training thus serves as an ultimate use case for demonstrating its support for complex, long-horizon tasks.

\subsubsection{Setup}
\paragraph{Models.}
\system{} organizes experiments and analysis. GLM-5.3 primarily serves as the controller and data teacher, while Qwen3.6-35B-A3B is the model trained and evaluated.

\paragraph{Initial state and iteration process.}
Starting from the base model, we perform seed supervised fine-tuning (SFT) on 1,345 tool-interaction trajectories from multiple sources to establish the R0 baseline, followed by five feedback-driven post-training rounds. Each round starts from evaluation of the current model and follows the cycle: \textbf{model evaluation $\rightarrow$ capability-gap diagnosis and data-pipeline revision $\rightarrow$ data generation, validation, and filtering $\rightarrow$ SFT $\rightarrow$ re-evaluation.}

\paragraph{Data quality and contamination control.}
All training data undergo BM25-based task-similarity filtering before training to control near-duplicate overlap between training and evaluation tasks.

\paragraph{Evaluation scope.}
We report results after each round on eight benchmarks: AutomationBench, BFCL v4, OfficeQA Pro (OfficeQA), Tau-3 Bench Banking, Harvey Legal Agent Bench, SkillsBench, GDPVal, and JobBench. GDPVal uses a normalized cumulative rubric score, separately from the Elo metric in Section~\ref{sec:benchmarks}. The aggregate is the arithmetic mean of the eight percentage scores, with each benchmark contributing up to 12.5 points. Further details appear in Appendix~\ref{sec:rsi-provenance}.

\subsubsection{Results}
\begin{figure}[tbp]
\centering
\includegraphics[width=\linewidth]{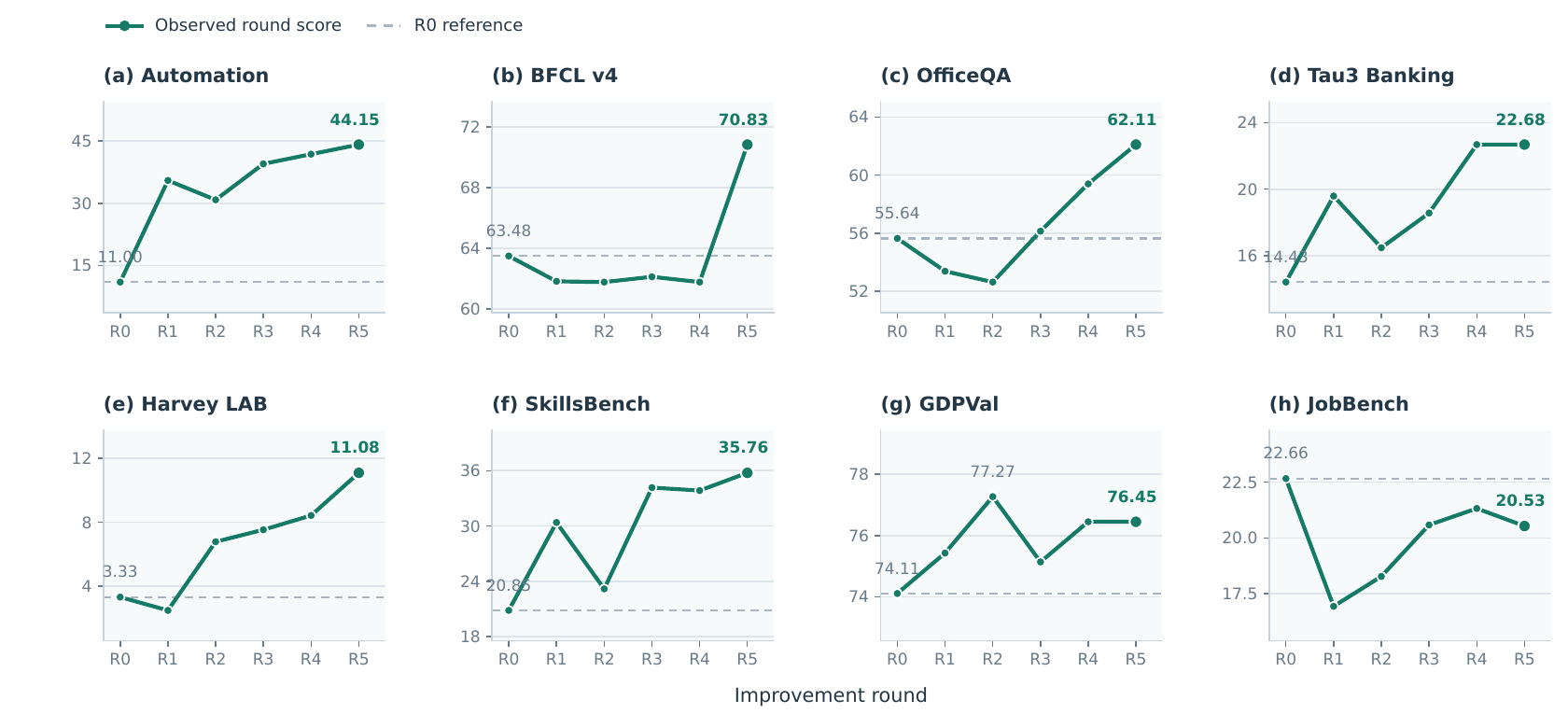}
\caption{Per-benchmark performance across R0--R5 in RSI-native automated post-training. The $2\times4$ panels show Automation, BFCL v4, OfficeQA, Tau3 Banking, Harvey LAB, SkillsBench, GDPVal, and JobBench. Dashed lines mark the R0 reference for each task. Labels identify the initial and final scores and any intermediate maximum distinct from the final score. The horizontal axis denotes iteration round.}
\label{fig:rsi-benchmarks}
\end{figure}

\begin{figure}[tbp]
\centering
\includegraphics[width=\linewidth]{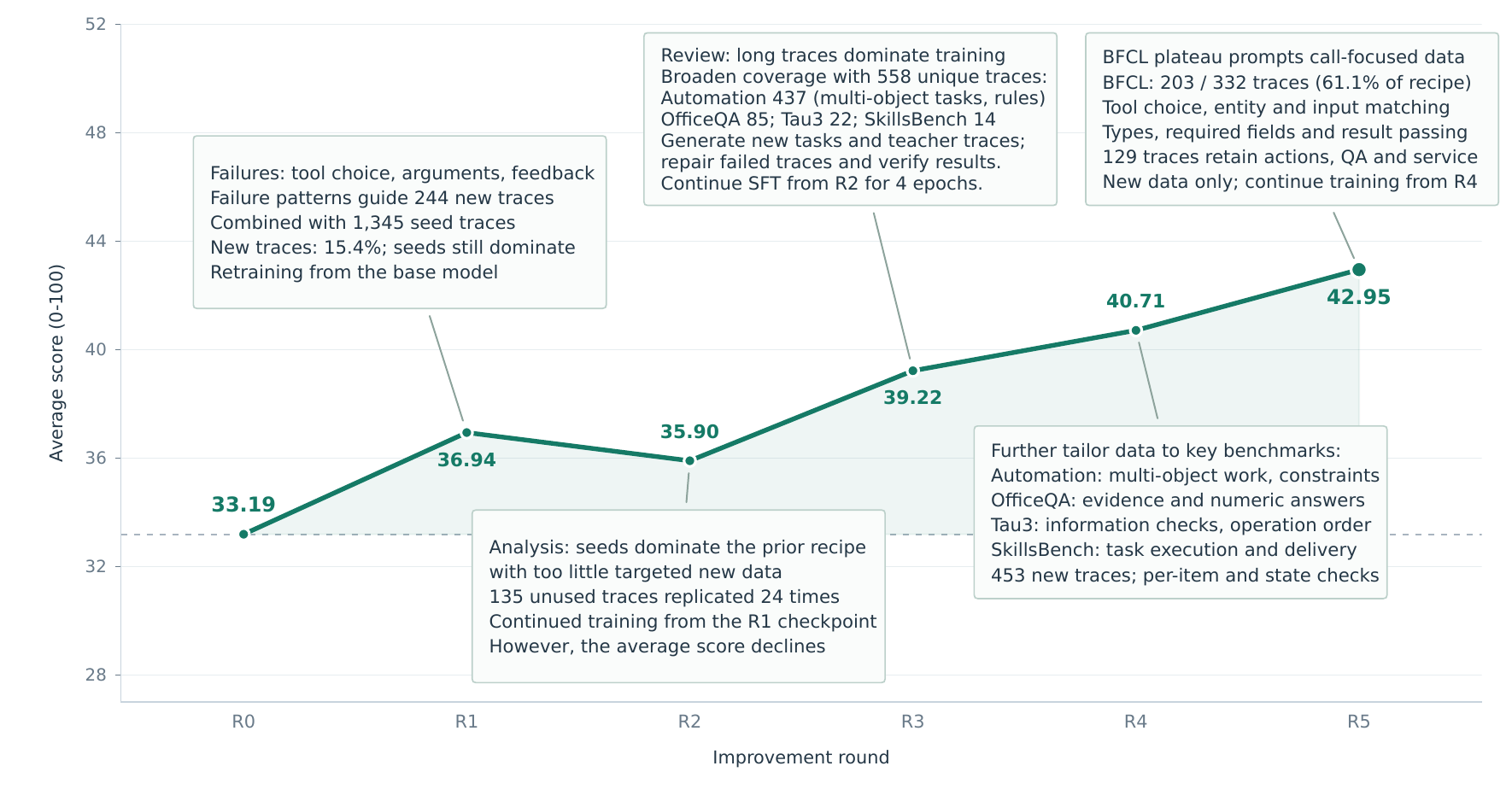}
\caption{Average score across eight benchmarks and key changes in RSI-native automated post-training. The curve connects observed averages at R0--R5; the dashed line marks the initial seed-SFT baseline. Each annotation integrates the corresponding round's checkpoint start, epochs, optimization steps, data recipe, and strategy changes. Sampling multipliers are reported separately from epochs. R0 and R1 start independently from the base model; R2--R5 continue from the preceding checkpoint. These changes describe concurrent interventions rather than isolated causal effects; round index does not represent elapsed time. Detailed training records appear in Appendix~\ref{sec:rsi-provenance}.}
\label{fig:rsi-progress}
\end{figure}

Across five post-training rounds, the model's average score on the eight benchmarks rises from 33.19 for the initial SFT model to 42.95, a gain of 9.76 points (Figure~\ref{fig:rsi-progress}). The average increases consecutively from R3 through R5. The final model exceeds the initial baseline on seven of eight benchmarks and achieves the highest score across the six rounds on six, including ties (Figure~\ref{fig:rsi-benchmarks}), demonstrating \system{}'s ability to sustain model improvement and broaden the gains from post-training.

Balancing multiple task capabilities is a central value of this result. Each round uses a single training dataset to update the same model, which must perform across all eight benchmarks; data selection and training strategies therefore need to accommodate different tool environments and behavioral requirements. Although later incremental supervision focuses on a subset of tasks, the final gains span function calling, complex tool execution, document question answering, and professional tasks, with seven benchmarks exceeding the baseline. This broad coverage demonstrates \system{}'s ability to balance targeted reinforcement with performance across tasks and provides positive evidence of cross-task generalization.

At the task level, Automation and SkillsBench show the largest gains of 33.15 and 14.91 points over R0, respectively, while Harvey LAB, Tau3 Banking, and OfficeQA show more sustained improvement. BFCL v4 gains 9.07 points from R4 to R5, consistent with the additional function-calling supervision in that round, whereas GDPVal and JobBench fluctuate overall. The best-performing round also differs by task: GDPVal peaks earlier, while Tau3 Banking matches its previous best in the final round and OfficeQA and BFCL v4 improve more substantially in later rounds. These differences reflect the changing trade-offs among tasks in joint post-training; the following analysis examines them through the evolution of the pipeline and training data.

\subsubsection{Analysis}
Across five rounds, \system{} continually uses evaluation feedback to identify capabilities to strengthen, revise the data recipe, and update the training strategy.

\paragraph{Identifying priorities from benchmark performance.}
The eight tasks do not respond uniformly to the same training round, and an increase in the overall average can coexist with stagnation on individual tasks. Per-benchmark results therefore provide a more specific basis for prioritizing data in the next round. Automation accounts for 78.3\% of the training trajectories at R3 and continues to receive additional data in the following two rounds. Its performance improves consecutively from R3 onward, rising from 39.49 to 44.15 at R5 and ultimately yielding the largest gain among the tasks. BFCL, by contrast, remains below the initial baseline throughout R1--R4. At R5, the data emphasis shifts toward function calling, which accounts for 61.1\% of the new trajectories, alongside a marked improvement on BFCL. The remaining 129 trajectories continue to cover complex operations, document question answering, and business interactions, retaining training support for other tasks while concentrating on function calling.

\paragraph{Improving demonstrations based on failure patterns.}
After identifying the tasks to strengthen, \system{} further adjusts demonstration content to address specific failed behaviors. R4 targets omissions in multi-item tasks, constraint violations, and missing verification after execution, requiring trajectories to demonstrate item-by-item handling, checks on operational constraints, and post-action state readback. Constraint adherence also becomes an acceptance criterion, leading to the exclusion of multiple Automation candidates at R5. For function calling, R5 further supplements demonstrations of entity and argument binding, parameter types, and required fields. These adjustments translate evaluation findings into concrete training requirements, with acceptance checks reducing the inclusion of erroneous demonstrations.

\paragraph{Planning training around training history and supervision distribution.}
To address R1's retraining from base and its high proportion of seed data, R2 selects the R1 checkpoint as its training starting point, stops replaying seed and previously trained data, and repeats 135 previously untrained trajectories 24 times. A review following the decline in average score from 36.94 to 35.90 finds that 42 trajectories contribute 84.2\% of supervised tokens, giving a small number of long trajectories a large share of the supervision. R3 retains incremental fine-tuning, expands the independent set to 558 trajectories, and trains from the R2 checkpoint for four epochs, using broader data coverage to reduce reliance on extensive repetition of a small corpus. The average score recovers to 39.22. Training adjustments thus extend from whether data have already been used to the coverage, supervision distribution, and repetition intensity of new data.

These adjustments demonstrate \system{}'s ability to accumulate evidence and revise experimental plans in complex, long-horizon research, providing empirical support for AI agents undertaking sustained model-development work.

\section{Conclusion}
\label{sec:conclusion}
We presented \system{}, a cowork harness for long-horizon, human--agent work. Its central design separates task strategy from execution management: Main selects and coordinates the work, while the runtime manages delegated execution, communication, shared application resources, persistent workspaces, semantic memory, and long-running Jobs. This organization allows agents and users to continue working on the same evolving artifacts while preserving the state and evidence needed across interruptions and context changes.

Across 12 benchmarks and six categories of application cases, \system{} achieves competitive performance relative to OpenHands, Codex, and DeepSeek Harness. In an RSI-native post-training study, it raises the average score of Qwen3.6-35B-A3B from 33.19 to 42.95 over five rounds and eight benchmarks. Although the gains vary across tasks, the results support a unified harness for complex productivity work, continued monitoring, and iterative model improvement.

\section{Author Contributions}
\label{sec:author-contributions}

\begingroup
\newcommand{\contribsymbol}[1]{{\normalfont\ensuremath{#1}}}
\newcommand{\contribmark}[1]{%
  \nobreak\hspace{0.05em}%
  \raisebox{0.55ex}{\fontsize{8}{8}\selectfont\contribsymbol{#1}}%
}
\newcommand{\contribequalmark}{\contribmark{\ast}}
\newcommand{\contribcorrespondingmark}{\contribmark{\dagger}}
\newcommand{\contribaffil}[1]{%
  \nobreak\hspace{0.05em}%
  \raisebox{0.55ex}{\fontsize{8}{8}\selectfont #1}%
}
\newcommand{\contribname}[1]{\mbox{#1}}

\system{} is a collaborative project. We summarize the contributors' primary roles below.

\subsection*{Core Contributors}
\contribname{Jiajie Jin}\contribaffil{1}\contribequalmark, \contribname{Guanting Dong}\contribaffil{1}\contribequalmark, \contribname{Chengjun Pan}\contribaffil{3}\contribequalmark, \contribname{Xin Deng}\contribaffil{2}, \contribname{Zhuohui Sheng}\contribaffil{2}, \contribname{Guoqiang Wang}, \contribname{Tao Gui}\contribaffil{2}\contribcorrespondingmark, \contribname{Yutao Zhu}\contribaffil{1}\contribcorrespondingmark.

\subsection*{Contributions by Area}
\begin{itemize}[leftmargin=1.15em,itemsep=7pt,topsep=2pt,parsep=0pt]

\item \textbf{General Capability Optimization:}
\contribname{Yujia Chen}\contribaffil{2}, \contribname{Shihong Yi}\contribaffil{2}, \contribname{Binze Hu}\contribaffil{2}, \contribname{Zhixiong Yang}\contribaffil{2}, \contribname{Zhichao Sun}\contribaffil{2}, \contribname{Rui Wang}\contribaffil{2}, \contribname{Han Han}\contribaffil{2}, \contribname{Mingyu Ma}\contribaffil{2}, \contribname{Junhao Huang}\contribaffil{4}, \contribname{Xiaotao Wang}\contribaffil{2}, \contribname{Guoqiang Wang}, \contribname{Huicheng Jiang}\contribaffil{2}.

\item \textbf{User Experience Optimization:}
\contribname{Yuyang Hu}\contribaffil{1}, \contribname{Xinyu Yang}\contribaffil{1}, \contribname{Wanghan Xu}\contribaffil{4}, \contribname{Yujia Chen}\contribaffil{2}, \contribname{Shihong Yi}\contribaffil{2}, \contribname{Guoqiang Wang}, \contribname{Yajie Yang}\contribaffil{2}.

\item \textbf{RSI Optimization:}
 \contribname{Guanting Dong}\contribaffil{1}, 
 \contribname{Jiajie Jin}\contribaffil{1},
 \contribname{Zhuohui Sheng}\contribaffil{2}, \contribname{Xin Deng}\contribaffil{2}, \contribname{Zhixiong Yang}\contribaffil{2}, \contribname{Han Han}\contribaffil{2}, \contribname{Mingyu Ma}\contribaffil{2}.

\item \textbf{Harness Development and Maintenance:}
\contribname{Jiajie Jin}\contribaffil{1}, \contribname{Chengjun Pan}\contribaffil{3}, \contribname{Yuyang Hu}\contribaffil{1}, \contribname{Zhuohui Sheng}\contribaffil{2}, \contribname{Guanting Dong}\contribaffil{1}.

\item \textbf{Discussion and Support:}
\contribname{Zhecan Wang}, \contribname{Honglin Guo}\contribaffil{2}, \contribname{Xiaoyu Xing}\contribaffil{2}, \contribname{Xiaoqian Zhang}\contribaffil{6}, \contribname{Yulun Zhu}\contribaffil{2}, \contribname{Xiaoxi Li}\contribaffil{1}, \contribname{Jiahang Lin}\contribaffil{2}, \contribname{Shaofan Liu}\contribaffil{2}, \contribname{Shihan Dou}\contribaffil{2}, \contribname{Zhiheng Xi}\contribaffil{2}, \contribname{Qi Zhang}\contribaffil{2}.
\end{itemize}

\subsection*{Project Mentorship}
\contribname{Tao Gui}\contribaffil{2}\contribcorrespondingmark, \contribname{Yu-Gang Jiang}\contribaffil{2}, \contribname{See-Kiong Ng}\contribaffil{5}, \contribname{Yang Zhang}\contribaffil{5}, \contribname{Tat-Seng Chua}\contribaffil{5}, \contribname{Yutao Zhu}\contribaffil{1}\contribcorrespondingmark, \contribname{Ji-Rong Wen}\contribaffil{1}, \contribname{Zhicheng Dou}\contribaffil{1}\contribcorrespondingmark.

\subsection*{Affiliations}
{
\noindent\contribaffil{1} Renmin University of China\\
\contribaffil{2} Fudan University\\
\contribaffil{3} Peking University\\
\contribaffil{4} Shanghai Jiao Tong University\\
\contribaffil{5} National University of Singapore\\
\contribaffil{6} East China Normal University.\par}

{
\contribsymbol{\ast}\enspace Equal contribution (co-first authors).\\
\contribsymbol{\dagger}\enspace Corresponding authors.\par}

\endgroup

\begingroup
\raggedright

\par
\endgroup
\FloatBarrier
\clearpage
\bibliographystyle{plainnat}
\bibliography{citation}
\FloatBarrier
\clearpage
\appendix
\section{Evaluation Details and Supplementary Evidence}
\label{sec:eval-details}

This appendix summarizes benchmark-specific evaluation settings, qualitative case evidence, and automated post-training run records.

\subsection{Benchmark Evaluation Settings}
\label{sec:benchmark-protocols}

\paragraph{SkillsBench~\citep{skillsbench2026}.}
We use the 79 text-based tasks from SkillsBench, excluding visual and media tasks, with an independent workspace and task container for each instance. We retain the official task environments and verifiers and report the mean task score.

\paragraph{$\tau^3$-Bench Banking~\citep{tauknowledge2026}.}
The evaluation contains 97 Banking Knowledge tasks, with BM25 retrieval, dense retrieval over 698 documents, and read-only shell access to the knowledge base. Each task allows up to 200 interaction steps, with a 1,800-second simulation limit and a 2,400-second outer task limit; we report task pass rate.

\paragraph{AutomationBench~\citep{automationbench2026}.}
We evaluate AutomationBench v1.0.6 with the null-type handling fix from PR~\href{https://github.com/zapier/AutomationBench/pull/13}{\#13}. Tasks are scored with the benchmark evaluator, and we report the all-pass rate.

\paragraph{JobBench~\citep{jobbench2026}.}
We use the main split of 65 tasks, with a 7,200-second task timeout. Deliverables receive task-level percentage scores from an LLM judge, averaged over all 65 tasks.

\paragraph{Harvey Legal Agent Bench~\citep{harveylab2026}.}
We use a fixed, proportionally sampled set of 120 legal tasks, retaining the official task documents, skills, and rubrics in isolated workspaces. Generated deliverables are graded criterion by criterion against the supplied rules and rubrics.

\paragraph{GDPVal~\citep{gdpval2025}.}
For each harness, we include all 220 scheduled tasks and report Elo against a fixed baseline of 1000, computed as $E=1000+400\log_{10}[(W+T/2)/(L+T/2)]$, where $W$, $L$, and $T$ are wins, losses, and ties. Tasks without a valid score count as losses, including candidate timeouts and call-gate exhaustion; thus $W+L+T=220$ for every harness.

\paragraph{OfficeQA Pro~\citep{officeqa2026}.}
We use the 133-question OfficeQA Pro subset and its parsed Treasury Bulletin corpus, with a fresh workspace for each question. Answers are scored by the native EvalScope/OfficeQA rules.

\paragraph{WorkspaceBench-Lite~\citep{workspacebench2026}.}
WorkspaceBench-Lite is built on the AgentCompass evaluation framework and uses the English Lite dataset. Each task is allocated 2 CPUs and 8 GiB of memory, with a two-hour candidate execution limit.

\paragraph{WideSearch~\citep{widesearch2025}.}
The task set comprises 200 questions, split equally between English and Chinese, with a shared Serper search and Jina page-reading backend. Answers are graded with the \texttt{format\_normalized\_v1} evaluator and an LLM judge.

\paragraph{BrowseComp~\citep{browsecomp2025}.}
Tasks use web search, page reading, and Python, with a maximum of 500 interaction steps. We use discard-all context management and report answer accuracy.

\paragraph{MLE-Bench-Lite~\citep{chan2025mlebench}.}
We use all 22 low-split competitions, with a five-hour agent budget, 4 CPUs, 32 GiB of memory, and one H200 GPU per task. The official grader scores saved CSV submissions, which are converted to Human Rank using the retained competition leaderboards and equally averaged across the 22 tasks; missing or invalid submissions receive zero.

\paragraph{MLS-Bench~\citep{mlsbench2026}.}
We evaluate the 30 Lite tasks with a five-hour agent budget, task-specific images, and task-specific H200 GPU allocations. The standard per-task allocation is 4 CPUs, with 32 GiB of memory for CPU-only and single-GPU tasks and 64 GiB for multi-GPU tasks. Only the declared algorithm files may be modified; official verifier rewards are scaled from 0--1 to 0--100 and equally averaged across tasks.

\paragraph{Domain averages.}
Each domain average equally weights its listed benchmarks; GDPVal contributes the tie-adjusted win rate $100(W+T/2)/(W+L+T)$, calculated directly from the exact counts rather than rounded Elo. The remaining benchmarks use their reported 0--100 scores, with rounding applied only after aggregation.

\subsection{Case Evidence and Inspection Protocol}
\label{sec:case-protocols}

Figure~\ref{fig:four-category-comparison} covers 24 cases and 72 artifacts across six categories. Comparisons use unretouched images, shared aligned CAD inspection views, and authored Blender materials and lighting.

\begin{figure}[!htbp]
\centering
\captionsetup{skip=3pt}
\includegraphics[width=0.75\linewidth]{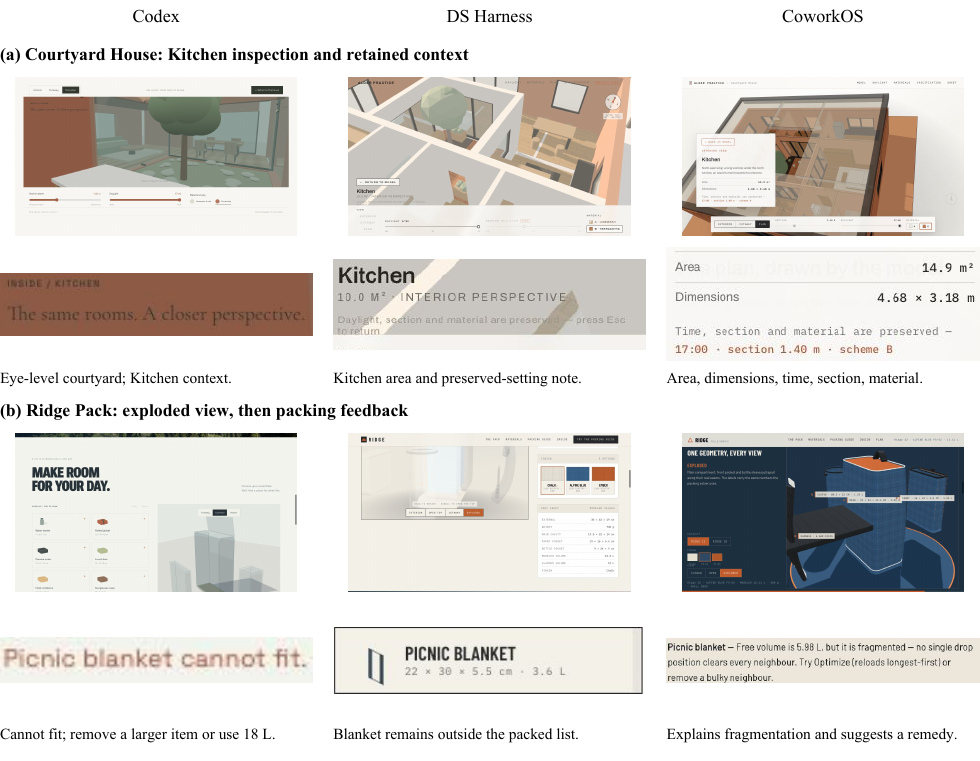}
\caption{Frontend website interaction evidence. Columns show Codex, DSH, and \system{}. (a) Kitchen inspection with room information. (b) Exploded pack views, followed by feedback from the recorded item-insertion sequence. Each panel pairs a scene view with a native interface crop; English notes below the crops summarize the observations.}
\label{fig:web-interaction-details}
\end{figure}

\paragraph{Frontend website evidence.}
Figure~\ref{fig:web-interaction-details}a compares Kitchen inspection in \emph{Courtyard House}. The reviewer adjusts section height, time, and material before selecting the room; all three implementations provide local inspection and retain the settings on return. Codex pairs an eye-level courtyard view with section, time, and material controls at the bottom, without listing area or dimensions in the room description. DSH labels the 10.0\,m$^2$ area and notes that daylight, section, and material are preserved. \system{}'s room card combines 14.9\,m$^2$, dimensions of 4.68$\times$3.18\,m, and the retained 17:00 time, 1.40\,m section, and material scheme B. The comparison concerns how object information and current settings are presented; retaining settings is shared by the three implementations.

Figure~\ref{fig:web-interaction-details}b shows feedback in \emph{Ridge Pack}. After attempts to insert the same eight named items in sequence, Codex reports that the picnic blanket cannot fit and suggests removing a larger item or using the 18\,L pack. DSH's recorded view lists compartment coordinates for packed items, with the blanket remaining unpacked; its transient rejection message was not captured, so the record does not establish whether it explains the cause. \system{}'s displayed message distinguishes spare volume from usable contiguous space: 5.98\,L remains, but fragmentation prevents any placement from clearing neighboring items. It also suggests the Optimize action or removal of a bulky neighbor. This supports a more specific rejection explanation than Codex's visible message. The implementations use different authored item dimensions, so packed-item counts and capacity values are not used to rank packing algorithms.

\begin{figure}[!htbp]
\centering
\includegraphics[width=0.88\linewidth]{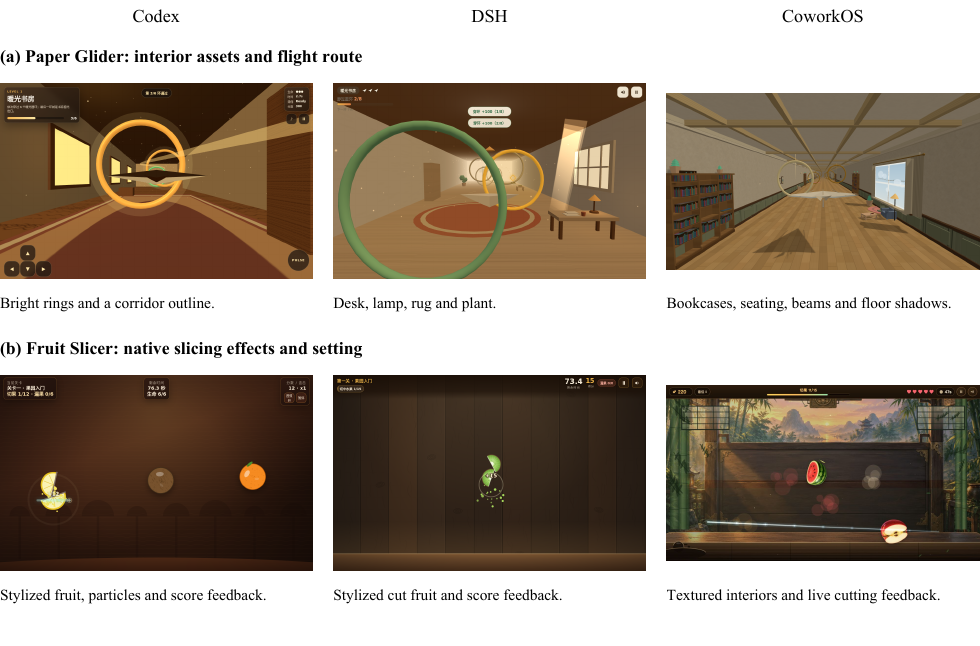}
\caption{Game scenes and interaction details. Columns show Codex, DSH, and \system{}. (a) Interiors and flight routes in \emph{Paper Glider}. (b) Slicing effects and environments in \emph{Fruit Slicer}, including a frame from pointer-driven play. Native views are scaled proportionally without retouching.}
\label{fig:game-feedback-details}
\end{figure}

\paragraph{Game scenes and interaction evidence.}
Figure~\ref{fig:game-feedback-details} expands the \emph{Paper Glider} and \emph{Fruit Slicer} comparisons. The following checks cover inputs and feedback across all three harnesses in four games.

\emph{Bubble Bastion.} Codex shows a cross-shaped splash, crate particles, and an increased crate-clearance counter; restarting clears this progress. DSH awards points after a bubble burst near a bush, presents a failure screen, and resets the score and enemy objective on retry. \system{} previews the cross-shaped danger area before the burst, repeats the objective in its pause panel, and restores the initial objective and hearts after a splash-related failure and retry.

\emph{Paper Glider.} All three respond to steering, display ring progress, and provide failure and retry feedback. Codex reports missed target rings, while DSH couples ring passage with point notifications. \system{}'s result dialog explains the missed-ring allowance; retry resets the progress. Figure~\ref{fig:game-feedback-details}a shows the scene differences: Codex foregrounds bright rings, windows, and a corridor outline; DSH adds a desk, lamp, rug, and plant. \system{} integrates book-filled shelves, curtains, seating, wall panels, ceiling beams, and floor shadows into a furnished study surrounding the flight route.

\emph{Racing.} Codex and DSH register steering, checkpoint progress, and nitro consumption; restart resets checkpoint and timing displays. DSH also shows the vehicle and checkpoints on a minimap and issues a wrong-way warning. In \system{}, throttle and steering change vehicle speed, heading, and distance to the next checkpoint. Pressing T after driving onto the shoulder returns the car to checkpoint~0 with an explicit message while retaining elapsed time; pressing R restarts the run and resets the timer.

\emph{Fruit Slicer.} Pointer drags increase cut counts and scores in all three, with visible fruit separation and juice effects. Codex shows combo and double-slice feedback; DSH increments a missed-fruit counter before presenting failure and retry. \system{} combines detailed fruit interiors, blade trails, and particles with bamboo, mountain scenery, and a wooden dojo frame (Figure~\ref{fig:game-feedback-details}b). Its HUD reaches the cut target and invites continued scoring; the observed run subsequently ends after excessive misses. Retry resets progress while retaining the best score. Displayed scores and times reflect each game's own rules and are not used to rank harness success rates.

\begin{figure}[!htbp]
\centering
\includegraphics[width=\linewidth]{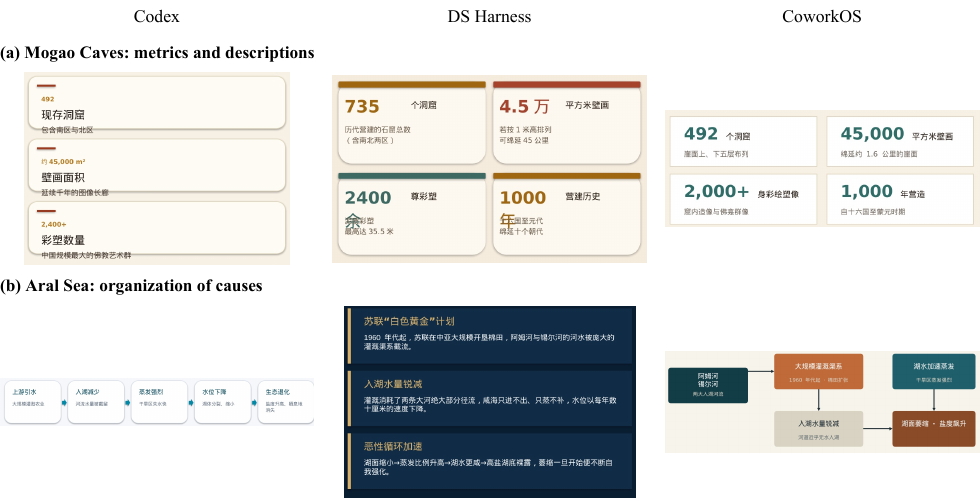}
\caption{Office content-page details. Columns show Codex, DeepSeek Harness (DSH), and \system{}. (a) Mogao Caves metric groups, from pages 2 / 3 / 2. (b) Aral Sea cause explanations, from pages 5 / 3 / 5. Rectangular crops retain each complete metric group or relation representation; native text and graphics are unchanged.}
\label{fig:office-content-details}
\end{figure}

\paragraph{Office content-page evidence.}
Figure~\ref{fig:office-content-details} preserves two sets of native content-page details used in the main comparison. In \emph{Mogao Caves} (a), the descriptions in Codex's three metric cards extend below their bottom borders. In DSH's lower two cards, the suffixes meaning ``more than'' and ``years'' wrap below the large numbers and overlap the descriptions. \system{} keeps each value, label, and explanation inside its card. The pages use different figures and wording; the comparison concerns text containment and metric grouping, not numerical correctness. In \emph{Aral Sea} (b), Codex's upper arrow chain sequentially connects upstream diversion, reduced inflow, evaporation, falling water levels, and ecological degradation. Its full page also separates human activities from natural conditions in two text groups below the chain. DSH organizes the process into three explanatory cards. \system{} separates reduced inflow and evaporation into two diagram branches entering a shared outcome. The distinction is whether the two displayed paths are separated directly in the relation diagram, not whether the baselines mention the relevant causes.

The Codex / DSH / \system{} page selections are 2 / 3 / 2 for Mogao metrics and 5 / 3 / 5 for Aral causes. Rectangular crops preserve complete metric groups, the arrow chain, the explanatory-card group, or the branched diagram. The roasting pages in \emph{Coffee Journey} provide a further comparison: \system{} page 6 links numbered photo annotations to descriptions of the roaster and cooling tray; Codex page 8 separates heating, moisture loss, cooling, and resting into four stage cards; DSH page 6 pairs stage descriptions with a time curve. The first emphasizes component-to-text correspondence, while the latter two organize the process. This does not establish broader topic coverage for \system{}. These content-page observations do not extend to \emph{Butterfly Life Cycle}, for which only covers were inspected.

\begin{figure}[!htbp]
\centering
\includegraphics[width=\linewidth]{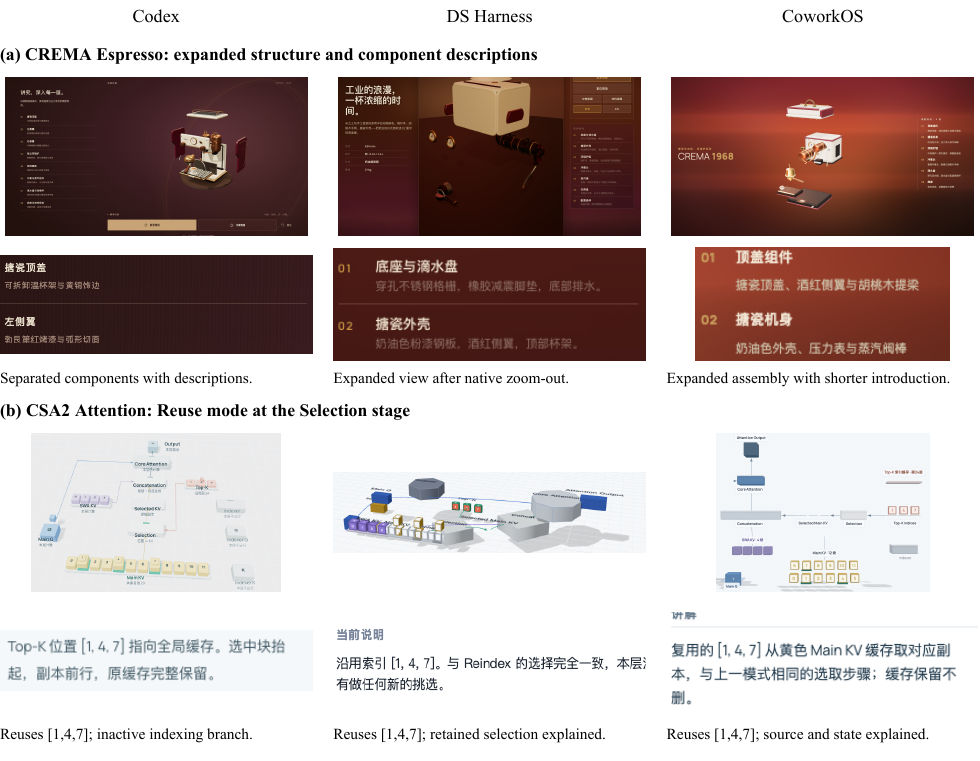}
\caption{Interactive visualization evidence. Columns show Codex, DSH, and \system{}. (a) Expanded CREMA Espresso assemblies and component-description crops; DSH uses its native zoom-out control. (b) CSA2 Attention in Reuse mode at Selection, with native explanation crops. English notes summarize the observations; native image content is unchanged.}
\label{fig:interactive-inspection-details}
\end{figure}

\paragraph{Interactive visualization evidence.}
In \emph{CREMA Espresso} (Figure~\ref{fig:interactive-inspection-details}a), native expansion changes component positions and presents descriptions in all three implementations. \system{} moves the camera back, reduces the title, and retracts the long introduction, keeping the separated lid, twin-boiler group, brew head, drip tray, and base within one view alongside six description groups. Codex separates the lid and side panels in different directions and keeps eight descriptions beside the assembly. DSH's initial expanded view clips the upper shell; native zoom-out brings it into view, but the opaque shell still occludes parts of the boilers. The figure uses this zoomed-out view. This comparison concerns framing and information organization in the expanded state; it does not imply that the baselines lack expansion or component descriptions.

For \emph{CSA2 Attention} (b), the reviewer switches from Full to Reindex and then Reuse at the Selection stage. Selected positions change from $[0,3,8]$ to $[1,4,7]$ and persist in Reuse. All three implementations display state changes, provenance colors, and stage explanations. \system{} uses front-facing labels to connect positions to yellow Main KV and coral Top-K, with text stating that corresponding entries are copied while the original cache remains. Codex likewise explains copy-forward behavior and cache preservation, and grays out inactive indexing; its in-scene position labels use a smaller oblique arrangement. DSH also explains index reuse, but its low-angle view crowds labels around Main Q, SWA KV, and Selection. These are legibility differences in the selected views, not validations of algorithmic equivalence or user comprehension. The advantage is also case-dependent: repeated reset, expansion, waiting, and rotation did not clearly expose separated internals in \system{}'s \emph{VESPER Speaker}, and Codex's \emph{Saturn Atlas} cutaway distinguished interior layers more clearly.

\begin{figure}[!htbp]
\centering
\includegraphics[width=\linewidth]{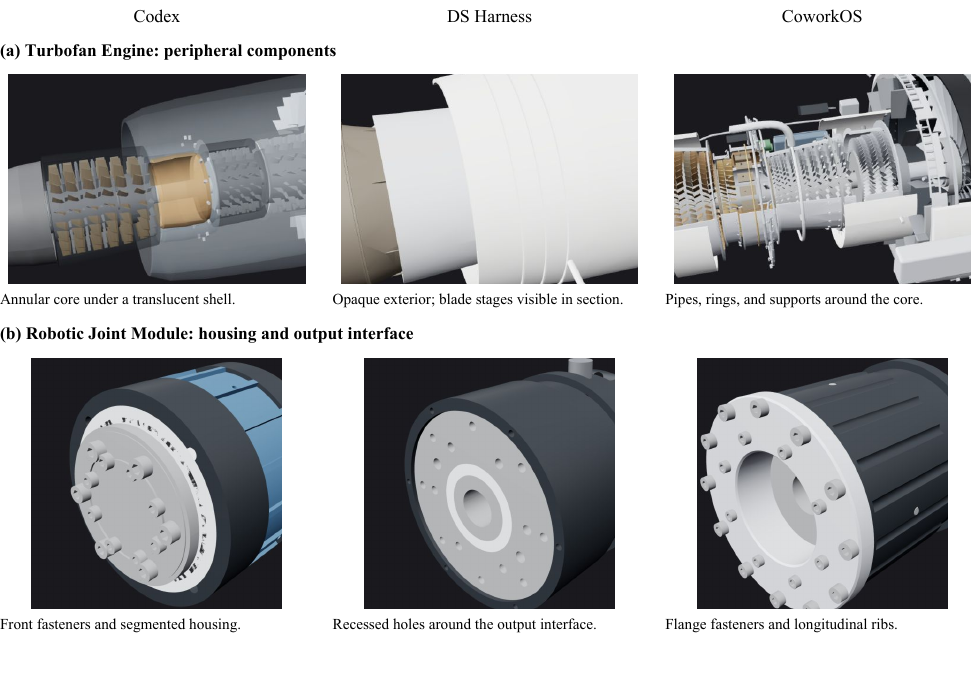}
\caption{Additional CAD evidence. Columns show Codex, DSH, and \system{}. (a) Peripheral regions of assembled turbofan engines. (b) Robotic-joint housing and output interfaces. Each row uses the same inspection view and crop rule after axis alignment. Authored cutaways, transparency, and opaque surfaces are retained. Unretouched crops demonstrate visible structural detail; the shared half-sections in Figure~\ref{fig:four-category-comparison}e expose the internals.}
\label{fig:cad-inspection-details}
\end{figure}

\paragraph{CAD structure and native-geometry checks.}
Figure~\ref{fig:cad-inspection-details} supplements the main half-sections with assembled details. In \emph{Turbofan Engine} (a), Codex shows axial blade groups and an annular core within a transparent casing, with relatively little peripheral equipment. DSH's opaque casing conceals the core at this angle, with annular bands and rods visible outside; the shared half-section in Figure~\ref{fig:four-category-comparison}e confirms its detailed multistage blade rows. \system{} distributes bent tubes, annular components, and block-like accessories around the core, supporting the main-text observation of richer peripheral assembly detail. In \emph{Robotic Joint Module} (b), both Codex and \system{} have housing ribs and populated flange fasteners; the full assembled \system{} view also shows a protruding side connection detail. DSH uses a smoother housing with flange holes and a central opening, while its section still exposes a layered interior. These differences concern the visual representation of peripheral structure and interfaces; casing occlusion is not used to infer missing internals.

All six STEP files were independently imported and transferred with OpenCascade / OCP 8.0.1.1.0 without source modification. Visible axial cavities in section also do not establish a continuous through-bore in the complete assembly. These observations support native readability, inspectability, and comparisons of visible structural detail; they do not establish parametric editability, interference-free fit, tolerances, manufacturability, or operating performance. Orbit, section, and wireframe controls belong to the shared inspection viewer.

\begin{figure}[!htbp]
\centering
\includegraphics[width=\linewidth]{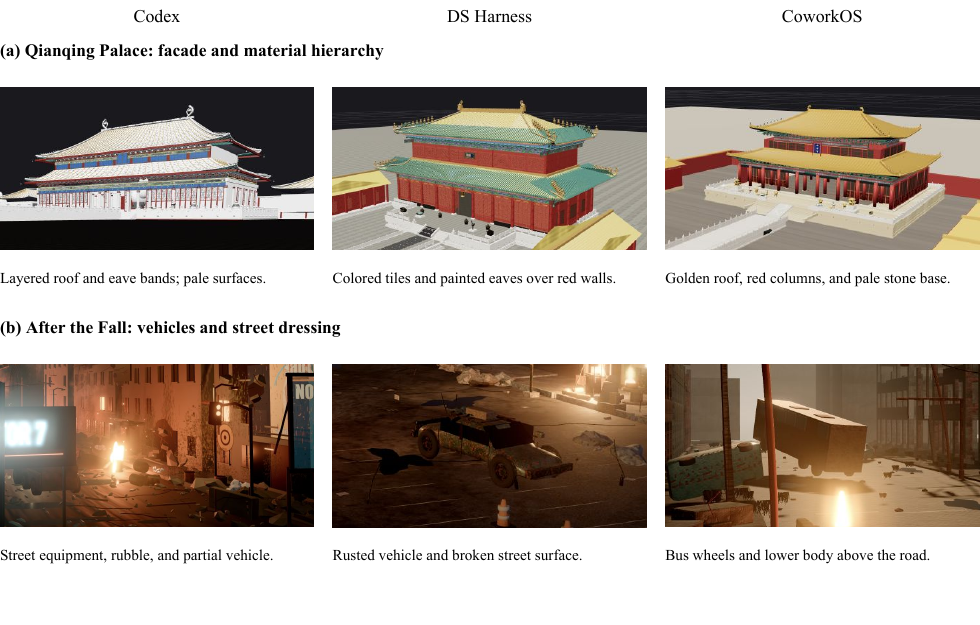}
\caption{Additional Blender scene evidence. Columns show Codex, DSH, and \system{}. (a) Corresponding main-hall facades after native orbit and zoom. (b) Vehicle and street-dressing crops from alternate views, retaining the visible bus--road gap in \system{}. Views retain the delivered models and presentation settings; they are not calibrated lighting comparisons. Only rectangular crops and proportional scaling are applied.}
\label{fig:blender-scene-details}
\end{figure}

\paragraph{Blender material hierarchy and scene organization.}
The main \emph{Qianqing Palace} views use corresponding principal-hall roof and terrace bounds; Figure~\ref{fig:blender-scene-details}a adds facades inspected through native orbit and zoom. Codex retains layered roofs, ridge ornaments, window divisions, a raised platform, and balustrades, with visible blue--green eave bands, although broad roof and wall surfaces appear pale. DSH clearly differentiates tile rows, painted eaves, and red door and window panels; its main facade has a less prominent projecting colonnade, and its side halls use simpler long roof and wall masses. \system{} separates golden roofs, red columns, lattice panels, painted eaves, and pale stonework into distinguishable components. The front colonnade, steps, and axial approach remain identifiable after changing viewpoint. These observations concern delivered materials and architectural legibility, not historical accuracy; mesh counts are not treated as quality scores.

In the main \emph{After the Fall} view, \system{} connects foreground vehicles, a mid-street bus, damaged multi-storey buildings, and distant shells into a continuous street. Cables span the road, and an emissive hotel sign anchors the right facade. Codex also has pronounced depth: its NO FUTURE and SECTOR 7 signs, dense rubble, and street equipment form a more crowded urban corridor. DSH combines a rusty vehicle, low-rise ruins, a bent street light, and an illuminated shelter in a more open setting. These differences concern spatial organization and asset composition, without establishing that \system{} leads on every scene detail. Alternate-view crops (b) retain vehicle and street details, including the visible gap between \system{}'s bus wheels or lower body and the road. Spatial depth does not establish correct prop support or collision. Views retain delivered materials and presentation settings rather than calibrated shared lighting. The evaluation uses GLB exports, which contain no animation tracks; viewpoint motion does not establish model animation.

\subsection{Automated Post-Training Run Records}
\label{sec:rsi-provenance}

\paragraph{Data contamination control.}
The initial seed data and all subsequent training data undergo BM25-based task-similarity filtering with a threshold of 0.8 to control near-duplicate overlap between training and evaluation tasks. Training-history deduplication separately checks which data have already been used by the parent model.

\paragraph{Scoring rules.}
All benchmarks follow their default scoring protocols. For GDPVal, the judge assigns partial credit according to task rubrics, and the reported score is $100\sum_{i\in V}s_i/\sum_{i\in V}m_i$, where $V$ is the set of validly judged tasks, $s_i$ is the actual score, and $m_i$ is the corresponding maximum. Actual scores and maxima are summed before division; this metric is neither Elo nor a reference win rate.

\paragraph{Complete results.}
Table~\ref{tab:rsi} reports all eight benchmark scores and the overall average for each of the six rounds. Scores use a 0--100 scale; the aggregate is their arithmetic mean, with each benchmark contributing up to 12.5 points. Bold and underlining mark the highest and second-highest distinct scores per row, with identical markings for ties; the change column is excluded from ranking.

\begin{table}[!htbp]
\centering
\setlength{\tabcolsep}{4pt}
\renewcommand{\arraystretch}{1.12}
\begin{tabular*}{\linewidth}{@{\extracolsep{\fill}}lrrrrrrr@{}}
\toprule
Benchmark & R0 & R1 & R2 & R3 & R4 & R5 & $\Delta$ R5--R0 \\
\midrule
Automation & 11.00 & 35.50 & 30.83 & 39.49 & \underline{41.83} & \textbf{44.15} & +33.15 \\
BFCL v4 & \underline{63.48} & 61.81 & 61.76 & 62.11 & 61.76 & \textbf{70.83} & +7.35 \\
OfficeQA & 55.64 & 53.38 & 52.63 & 56.14 & \underline{59.40} & \textbf{62.11} & +6.47 \\
$\tau^3$-Banking & 14.43 & \underline{19.59} & 16.49 & 18.57 & \textbf{22.68} & \textbf{22.68} & +8.25 \\
Harvey LAB & 3.33 & 2.50 & 6.78 & 7.53 & \underline{8.42} & \textbf{11.08} & +7.75 \\
SkillsBench & 20.85 & 30.38 & 23.18 & \underline{34.16} & 33.84 & \textbf{35.76} & +14.91 \\
GDPVal & 74.11 & 75.43 & \textbf{77.27} & 75.14 & \underline{76.45} & \underline{76.45} & +2.34 \\
JobBench & \textbf{22.66} & 16.93 & 18.27 & 20.58 & \underline{21.32} & 20.53 & $-2.13$ \\
\midrule
Average / 100 & 33.19 & 36.94 & 35.90 & 39.22 & \underline{40.71} & \textbf{42.95} & +9.76 \\
\bottomrule
\end{tabular*}
\caption{Complete results for AI-automated post-training across R0--R5 and the final change relative to the initial baseline.}
\label{tab:rsi}
\end{table}

\paragraph{Training lineage and data.}
R0 is the baseline obtained through seed SFT from the base model. R1 also trains again from base, rather than adding another round of seed training to R0. From R2 onward, training-history checks select trajectories not previously used by the parent model, separately recording previously accepted but unused data and newly generated data. Independent trajectory counts differ from the number of sample presentations after repetition.

\begin{center}
\small
\setlength{\tabcolsep}{4pt}
\renewcommand{\arraystretch}{1.15}
\begin{tabularx}{\linewidth}{@{}llYclr@{}}
\toprule
Round & Start & Independent trajectories and composition & Epochs & \shortstack[l]{Sampling\\multiplier} & Steps \\
\midrule
R0 & base & 1,345 seeds from five sources & 1 & $1\times$ & 66 \\
R1 & base & 1,345 seeds + 244 additions & 1 & $1\times$ & 69 \\
R2 & R1 & 135 first-use: 77 previously retained + 58 newly generated & 1 & \shortstack[l]{New $24\times$\\Old $1\times$} & 10 \\
R3 & R2 & 558 first-use: AB 437, OQ 85, $\tau^3$-22, SB 14 & 4 & \shortstack[l]{New $24\times$\\Old $1\times$} & 24 \\
R4 & R3 & 453 first-use across AB, OQ, Tau3, SB & 4 & \shortstack[l]{New $24\times$\\Old $1\times$} & 22 \\
R5 & R4 & 332 new: BFCL 203, AB 84, OQ 26, $\tau^3$-17, JB 2 & 4 & \shortstack[l]{New $24\times$\\Old $1\times$} & 20 \\
\bottomrule
\end{tabularx}
\end{center}

The five seed sources for R0 contain 320 open-source BC trajectories, 24 self-built BC trajectories, 261 DSQA trajectories, and two sleep-filtered WS sets of 627 and 113 trajectories.

\end{document}